\documentclass[journal]{IEEEtran}
\usepackage{amsmath,amssymb}
\usepackage{booktabs}
\usepackage{array}
\usepackage{capt-of}
\usepackage{placeins}
\newsavebox{\ladderbox}
\usepackage{graphicx}
\usepackage{xcolor}
\usepackage{tikz}
\usetikzlibrary{arrows.meta,positioning,calc,fit,backgrounds}
\usepackage{pgfplots}
\pgfplotsset{compat=1.17}
\usepgfplotslibrary{groupplots}
\usepackage{cite}
\usepackage[hidelinks]{hyperref}
\graphicspath{{figs/}}

\newcommand{\X}{\mathbf{X}}

\newcommand{\hd}[1]{\begin{tabular}[b]{@{}c@{}}#1\end{tabular}}

\definecolor{lane-s}{RGB}{74,85,104}
\definecolor{lane-g}{RGB}{31,111,139}
\definecolor{lane-p}{RGB}{140,88,32}
\definecolor{fill-s}{RGB}{231,234,239}
\definecolor{fill-g}{RGB}{222,237,242}
\definecolor{fill-p}{RGB}{246,231,220}

\definecolor{rewrite}{RGB}{160,30,30}
\definecolor{rewrite-edited}{RGB}{190,110,20}

\begin{document}

\title{Screw Attention: Rigid-Body Algebra Inside a Transformer}

\author{Aly Magassouba%
\thanks{A. Magassouba is with the School of Computer Science, University of Nottingham,
Nottingham, United Kingdom (e-mail: aly.magassouba@nottingham.ac.uk). This work is licensed
under a Creative Commons Attribution 4.0 International (CC BY 4.0) licence.}}

\markboth{Screw Attention}{Screw Attention}

\maketitle

\begin{abstract}
Learned manipulation policies rediscover from data the spatial relations that rigid-body mechanics supplies in closed form, which leaves them fragile to geometric change. We present Screw Attention, a transformer layer in which the relation between two bodies is a spatial transform rather than a graph edge. Each pair of tokens carries the relative pose and, for robot joints, the joint screw. Messages are transported along this relation into the receiver's frame, while the attention scores see only frame-invariant quantities. By construction, the messages are equivariant to an independent change of frame at every token, and a single layer can express the velocity recursion of rigid-body mechanics. On LIBERO-Spatial, a policy of 16k parameters trained from object poses alone reaches 97.3\% success, above graph, transformer and flat networks of the same size and a flat network with 27 times more parameters. Ablations show that the gain comes from transporting the correct relations, and that the structure pays most where the task requires relations between frames that nothing else supplies. The equivariance makes the policy robust to how the robot is described, where every other learned network collapses under a change of frame convention. Furthermore, the policy tolerates pose noise and calibration errors at least as well as an analytic controller. Used as a gated residual on an analytic controller, it also improves a contact-rich insertion task. Code and trained policies will be released.
\end{abstract}


\section{Introduction}
\label{sec:intro}

\IEEEPARstart{L}{earned} policies have become one of the main approaches to robot manipulation,
using scale as their principal lever. For instance, vision-language-action models with billions of
parameters, trained on large corpora of demonstrations, now perform multi-step tasks from
raw images and natural language. However, recent robustness analyses expose a sharp weakness:
displacing the manipulated objects, or altering the robot's initial configuration, degrades the
accuracy of such policies drastically~\cite{liberoplus2025,liberopro2025}. These failures arise
in the representation of spatial relations as well as in perception.

Yet those relations are exactly what rigid-body mechanics supplies in closed form. The kinematic model specifies how motion at one joint propagates to distant links, how a force applied at the gripper is distributed across the actuators, and how the gripper pose relates to any commanded frame. These relations are available a priori, long before any data are collected. Classical controllers are built on this structure, which is why such controllers remain compact, predictable, and reliable within the regimes they are designed to cover. The limitations of this approach are equally well-defined: analytic methods cannot easily express contact, friction, deformation, uncertain payload properties, or any task whose objective is not a fixed set-point. In this respect, analytic methods and learned policies form two complementary paradigms. One is given the geometry but cannot represent what the model omits; the other can represent what the model omits but must rediscover the geometry.

Several lines of work have therefore reintroduced structure into learned policies, although each retains only part of it. Embodiment-aware policy
transformers~\cite{gupta2022metamorph,sferrazza2024body,morphformer2026} encode the robot's
connectivity as an abstract graph, through per-limb descriptors or by biasing attention toward links that are neighbours in the
kinematic tree. Equivariant manipulation policies~\cite{ryu2023diffusionedf,equact2026,savla2026}
preserve the geometry of the scene, so that behaviour is consistent when the scene is
transformed, but they represent the robot only through its end-effector pose.
Kinematics-structured policies~\cite{xmop2024,rodrinet2025} do carry the poses or transforms of
individual links, yet process them with learned matrices, discarding the very structure that made those poses worth carrying. In all cases, the relation between two bodies is either abstracted into a graph or reduced to coordinates that a learned map is free to reinterpret.

\begin{figure}[!t]
\centering
\resizebox{\columnwidth}{!}{%
\begin{tikzpicture}[
  x=1cm, y=1cm, font=\large,
  lk/.style={line width=3pt, draw=lane-s, line cap=round, line join=round},
  jt/.style={circle, draw=lane-s, fill=white, line width=1.1pt, inner sep=0pt,
             minimum size=6pt},
  tok/.style={circle, draw=lane-s, fill=fill-s, line width=1.1pt,
              inner sep=0.5pt, minimum size=17pt, font=\normalsize},
  edge/.style={draw=lane-s!55, line width=1.6pt},
  pose/.style={-{Stealth[length=2.2mm]}, draw=lane-p, densely dashed,
               line width=1.3pt},
  msg/.style={-{Stealth[length=2.4mm]}, line width=1.5pt},
  spin/.style={-{Stealth[length=1.6mm]}, line width=1.1pt, draw=lane-g},
  ax/.style={-{Stealth[length=1.6mm]}, line width=1.2pt,
             preaction={draw=white, line width=3pt}},
  ttl/.style={font=\large\bfseries, align=center},
  key/.style={anchor=west, text=lane-s},
  arm/.pic={
    \fill[lane-s!25] (-0.85,0.09) rectangle (0.85,0.38);
    \draw[lane-s!45] (-1.05,0.09) -- (4.95,0.09);
    \draw[lk] (0,0.38) -- (0.18,2.20) -- (2.05,3.05) -- (3.25,1.65);
    \draw[lk, line width=1.9pt] (3.25,1.65) -- (3.75,1.10);
    \node[jt] at (0.18,2.20) {}; \node[jt] at (2.05,3.05) {};
    \node[jt] at (3.25,1.65) {};
  }]

\newcommand{\frameat}[3]{%
  \draw[ax, draw=lane-p] (#1,#2) -- ++(#3:0.62);
  \draw[ax, draw=lane-g] (#1,#2) -- ++(#3+90:0.62);
  \fill[lane-s] (#1,#2) circle (2.2pt);}
\newcommand{\neuron}[2]{%
  \draw[lane-s, fill=white, line width=1pt] (#1,#2) circle (0.09);}
\newcommand{\screwat}[4]{
  \begin{scope}[shift={(#1,#2)}, rotate=#3, scale=#4]
    \draw[lane-g, line width=1.3pt, -{Stealth[length=1.9mm]}]
         (-0.42,0) -- (0.50,0);
    \draw[lane-g, line width=1.1pt, domain=0:15.71, samples=140, smooth,
          variable=\t]
         plot ({-0.34+0.043*\t+0.06*sin(\t r)}, {0.17*cos(\t r)});
  \end{scope}}
\newcommand{\flattok}[2]{%
  \draw[thick, draw=lane-s, fill=fill-s, rounded corners=1.5pt]
       (#1-0.5,-2.25) rectangle (#1+0.5,-1.05);
  \foreach \yy in {-1.30,-1.53,-1.76,-1.99} \neuron{#1}{\yy}
  \node[text=lane-s] at (#1,-2.60) {#2};}
\newcommand{\screwtok}[3]{%
  \draw[thick, draw=#3, fill=fill-s, rounded corners=1.5pt]
       (#1-0.5,-1.55) rectangle (#1+0.5,-1.05);
  \neuron{#1-0.18}{-1.30} \neuron{#1+0.18}{-1.30}
  \draw[thick, draw=#3, fill=fill-g, rounded corners=1.5pt]
       (#1-0.5,-2.25) rectangle (#1+0.5,-1.55);
  \screwat{#1}{-1.90}{0}{1.0}
  \node[text=lane-s] at (#1,-2.60) {#2};}

\node[ttl, text=lane-s] at (1.95,4.35)
  {(a) Embodiment-Aware\\ Transformer};
\pic at (0,0) {arm};
\draw[thick, draw=lane-s!60, fill=lane-s!10] (3.95,0.09) rectangle (4.70,0.80);
\node[tok] (a1) at (0.09,1.29) {1};
\node[tok] (aj) at (1.12,2.63) {$j$};
\node[tok] (ai) at (2.65,2.35) {$i$};
\draw[edge] (a1) -- (aj);
\draw[edge] (aj) -- (ai);
\draw[thick, draw=lane-s, fill=lane-s!55, rounded corners=1pt]
     (4.20,3.22) rectangle (4.90,3.62);
\fill[lane-s] (4.43,3.22) -- (4.67,3.22) -- (4.75,3.02) -- (4.35,3.02) -- cycle;
\draw[lane-s!45, densely dotted, thick] (4.38,3.02) -- (3.95,0.80);
\draw[lane-s!45, densely dotted, thick] (4.72,3.02) -- (4.70,0.80);
\draw[msg, draw=lane-s!60, rounded corners=4pt]
     (4.90,3.42) -- (5.35,3.42) -- (5.35,-1.65) -- (4.90,-1.65);
\flattok{0.00}{1}
\flattok{1.35}{$j$}
\flattok{2.70}{$i$}
\draw[thick, draw=lane-s, fill=white, rounded corners=1.5pt]
     (3.85,-2.25) rectangle (4.85,-1.05);
\fill[lane-s!15] (3.90,-2.20) rectangle (4.80,-1.85);
\draw[thick, draw=lane-s!60, fill=lane-s!10] (4.15,-1.85) rectangle (4.55,-1.45);
\node[text=lane-s] at (4.35,-2.60) {image};
\draw[msg, draw=lane-s] (1.45,-1.00) to[bend left=60]
     node[midway, above=4pt, text=lane-s] {$\mathbf{W} \mathbf{h}_j$} (2.60,-1.00);
\node[key] at (-1.0,-3.10) {\textbf{token $\mathbf{h}_j$:} a vector of neurons};
\node[key] at (-1.0,-3.58) {\textbf{pair $(i,j)$:} a joint, or none};
\node[key] at (-1.0,-4.06) {\textbf{message:} a learned vector};
\node[key] at (-1.0,-4.54) {\textbf{scene:} image features};

\draw[lane-s!30, line width=0.8pt] (5.85,-4.66) -- (5.85,4.62);

\begin{scope}[shift={(7.35,0)}]
  \node[ttl, text=lane-g] at (1.95,4.35) {(b) Screw Attention\\ (Ours)};
  \pic at (0,0) {arm};
  \draw[spin] (0.18,2.20) ++(-0.26,-0.10) arc (200:440:0.26);
  \node[text=lane-g, anchor=east] at (-0.12,2.52) {$\mathbf{S}_j$};
  \draw[thick, draw=lane-p, fill=fill-p] (3.95,0.09) rectangle (4.70,0.80);
  \draw[thick, draw=lane-s, fill=lane-s!55, rounded corners=1pt]
       (4.20,3.22) rectangle (4.90,3.62);
  \fill[lane-s] (4.43,3.22) -- (4.67,3.22) -- (4.75,3.02) -- (4.35,3.02) -- cycle;
  \draw[lane-s!45, densely dotted, thick] (4.38,3.02) -- (3.95,0.80);
  \draw[lane-s!45, densely dotted, thick] (4.72,3.02) -- (4.70,0.80);
  \draw[msg, draw=lane-p!75, rounded corners=4pt]
       (4.90,3.42) -- (5.35,3.42) -- (5.35,0.45)
       node[midway, right=1pt, text=lane-p] {pose} -- (4.78,0.45);
  \frameat{0.09}{1.29}{-30}
  \frameat{1.12}{2.63}{70}
  \frameat{2.65}{2.35}{10}
  \frameat{4.33}{0.45}{35}
  \node[text=lane-s] at (-0.25,1.60) {1};
  \node[text=lane-s] at (0.80,2.28) {$j$};
  \node[text=lane-s] at (3.05,2.78) {$i$};
  \node[text=lane-s] at (4.85,0.95) {$o$};
  \draw[pose, shorten <=4pt, shorten >=5pt] (1.12,2.63) to[bend right=45]
      node[midway, below=0pt, text=lane-p] {$\X_{ij}$} (2.65,2.35);
  \draw[pose, shorten <=4pt, shorten >=5pt] (4.33,0.45) to[bend right=30]
      node[midway, right=2pt, text=lane-p] {$\X_{io}$} (2.65,2.35);
  \screwtok{0.00}{1}{lane-s}
  \screwtok{1.35}{$j$}{lane-s}
  \screwtok{2.70}{$i$}{lane-s}
  \screwtok{4.35}{$o$}{lane-p}
  \draw[msg, draw=lane-p] (1.45,-1.00) to[bend left=60]
       node[midway, above=4pt, text=lane-p] {$\X_{ij}\, \mathbf{W} \mathbf{g}_j$} (2.60,-1.00);
  \screwat{2.025}{-0.71}{0}{0.85}
  \node[key] at (-1.0,-3.10) {\textbf{token $j$:} neurons $\mathbf{s}_j$, screws $\mathbf{g}_j$};
  \node[key] at (-1.0,-3.58) {\textbf{pair $(i,j)$:} pose $\X_{ij}$, screw $\mathbf{S}_j$};
  \node[key] at (-1.0,-4.06) {\textbf{message:} a screw, transported};
  \node[key] at (-1.0,-4.54) {\textbf{scene:} a token with a pose};
\end{scope}
\end{tikzpicture}}
\vspace{-1.2\baselineskip}
\caption{(a) An embodiment-aware transformer with graph edges and (b) Screw Attention, where
tokens are related by their pose $\X_{ij}$ and joint screw $\mathbf{S}_j$, along which
messages are transported; the choice of frames does not affect the output. $\mathbf{W}$ is a
learned matrix: in (a) it maps the token vector $\mathbf{h}_j$; in (b) it mixes the screw
channels of $\mathbf{g}_j$, and $\X_{ij}$ transports the result into the frame of $i$. An
object carries a pose but no joint screw.}
\label{fig:teaser}
\end{figure}
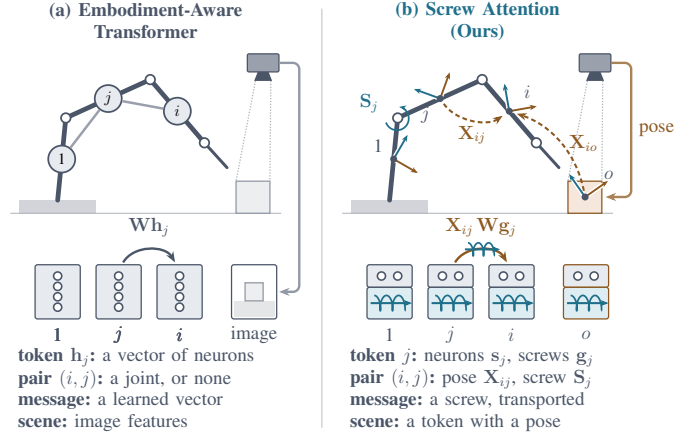

That relation is physical and has explicit content: the distance between two bodies, their relative orientation, and, when a joint is present, the axis of rotation or translation. Robotics provides a concise formalism for exactly these notions, the spatial or screw algebra of rigid-body
mechanics~\cite{featherstone2008,murray1994,lynch2017}. In this formalism such a relation is a single
object and composing two of them determines directly how motion of one body induces motion of another. A policy given these relations is therefore not required to learn kinematics from scratch, since the kinematic structure is already encoded in the relations themselves.

Our aim is an architecture that uses this kinematic
structure directly instead of relearning it. We adopt a transformer~\cite{vaswani2017attention} whose
tokens exchange screws (the velocities and forces of rigid-body mechanics) in addition to
feature vectors. As illustrated in Fig.~\ref{fig:teaser}, in our architecture, each input token is a body with a pose, which may be a robot link or an object in
the scene. Standard transformers relate tokens only through a learned compatibility between their features. In our approach, each pair of tokens also carries a given term that represents the physical relation between the two bodies, read from the kinematic model or, for objects, from perception. This addition enables two architectural shifts.

First, any message sent from one token to another is transformed into the receiving body's own
frame. Quantities defined at different locations cannot be combined until they are expressed in
a common frame, so the message must be transformed before the receiving token can use it.
Second, the attention weights access the relation only through frame-invariant quantities (e.g.,
distances, joint types, adjacency), which enter as an attention bias. A bias on the
scores only reweights the tokens, which a transformer can in principle learn unaided, whereas transporting a quantity
between coordinate frames is an operation it cannot express on its own. The physics therefore
enters through the messages.

Three properties follow from this construction. (i) The recursion that propagates velocities along the kinematic chain lies within what the layer can express, so the policy need not learn mechanics from data and can spend its capacity on the phenomena that mechanics omits. (ii) The geometry is
combined only through operations admitted by the spatial algebra, and these operations commute with changes of coordinate frame. As a result, the transported messages transform consistently when the frame attached to a link is changed, and the command does not depend on that choice, which is a modelling convention with no physical meaning. (iii) Because the readout emits the same command as an analytic controller, a joint torque or an end-effector twist, the layer can be used alone or on top of such a controller as a residual. Closing a gate recovers the controller exactly, so the balance between analytic and learned control is a choice made per task rather than fixed by the architecture.

In robotics, every link carries its own frame. The choice of where that frame is attached is a convention that varies across description formats~\cite{denavit1955kinematic} and sometimes even across descriptions of the same robot~\cite{tola2024urdf}. A policy whose behaviour depends on this convention can fail on a robot that is physically identical but described differently. Existing equivariant architectures~\cite{brehmer2023gatr,savla2026} are equivariant only to a single, global change of frame applied to the entire scene. By contrast, our layer is equivariant to an independent change of frame on every link, which is the freedom a robot description actually has. 

Our contributions are as follows.

\begin{enumerate}
 \item \textbf{An embodiment-aware attention layer grounded in rigid-body mechanics.} Tokens are related through the physical relation between bodies rather than an abstract graph, and messages are transported into the receiver's coordinate frame. As a result, the computation of how motion propagates along the robot can be expressed by the layer instead of being learned from data (Section~\ref{sec:representability}). Moreover, the layer can be used on its own or as a residual on top of an analytic controller.

 \item \textbf{An extensive empirical study across diverse tasks and a standard benchmark.} We evaluate our architecture on three simulated manipulation tasks and on LIBERO-Spatial, comparing it against mechanism-matched baselines and ablations. The structured policy outperforms graph, transformer and flat networks of the same size on LIBERO-Spatial, and even surpasses a flat network with $27\times$ more parameters. Under changes to the per-link frame convention, it is the only learned network on LIBERO-Spatial whose success is unchanged, whereas all learned baselines collapse to 0.0--2.3\%. Used on top of an analytic controller on a contact-rich insertion task, it raises the success rate by 17.3 points. These results show that the structure contributes capability (Sections~\ref{sec:pays}, \ref{sec:doesnot} and \ref{sec:libero}).

 \item \textbf{A hypothesis for when the structure is decisive}. Across the tasks studied, we identify a simple condition under which the geometry affects performance: the structure is decisive when the task requires reasoning over relations between coordinate frames that are not otherwise supplied by an analytic controller. When this condition is met, correct relations yield large gains over incorrect ones. When it is not, the correctness of the relations makes no measurable difference. This hypothesis is consistent with all results reported here (Section~\ref{sec:doesnot}, Appendix~\ref{app:nulls}).
\end{enumerate}

\section{Related Work}
\label{sec:related}

Four strands of prior work are directly relevant to this study: policy transformers that encode the robot's structure, equivariant policies that preserve scene geometry, architectures that represent the robot's kinematics explicitly, and methods that introduce physics through the training objective instead of the network. Each preserves a different aspect of the physical structure available to a manipulation policy.

\textbf{Embodiment-aware policy transformers.} This strand aims to exploit the robot's structure in a policy. MetaMorph~\cite{gupta2022metamorph}
assigns a token to each limb and conditions it on per-limb morphology descriptors. Body
Transformer~\cite{sferrazza2024body} restricts attention using a mask derived from the body
graph, and morphology-embedded transformers~\cite{morphformer2026} add a topology-aware
bias to the attention scores, with per-joint descriptors as ordinary features. In all cases, the kinematic tree enters only as connectivity: which parts exist and which are connected. The physical content of the relation between two links, how far apart they are and about which
axis one moves relative to the other, is however left for the network to infer. The scene is likewise represented as flat features rather than as additional bodies with poses.

\textbf{Equivariant manipulation policies.} Whereas the first strand structures the robot, this strand structures the scene. Motivated by data efficiency and by generalisation to new object placements, Diffusion-EDFs~\cite{ryu2023diffusionedf}, RiEMann~\cite{gao2024riemann},
EquiBot~\cite{yang2024equibot}, EquAct~\cite{equact2026} and SAVLA~\cite{savla2026} build policies whose output transforms consistently when the scene is rotated or translated, and general-purpose architectures such as the geometric-algebra transformer (GATr)~\cite{brehmer2023gatr} provide the underlying machinery. These methods obtain a strong geometric representation of the objects, but they represent the robot only through its end-effector pose. SAVLA is the closest to our layer in construction, carrying typed scalar and vector channels through its action head, yet it has no per-link tokens, no relative transforms between them and no joint motion subspaces.

Our layer differs from these architectures in its aim and in its symmetry. Their defining property is equivariance to one transformation of the whole scene, with every feature expressed in one common frame. In Screw Attention, invariance is a consequence of building the layer on the screw algebra of rigid-body mechanics rather than its aim, and the same construction reproduces classical computations of robot mechanics. Its symmetry group is also larger: the global frame change handled by existing equivariant architectures is the special case in which every token's frame is transformed in the same way (Appendix~\ref{app:gatr}). Finally, the first two strands have been geometric about different halves of the problem, the robot's structure and the scene's geometry. Our layer applies the same typing to the robot's own bodies and to the objects of the scene, within a single token space.

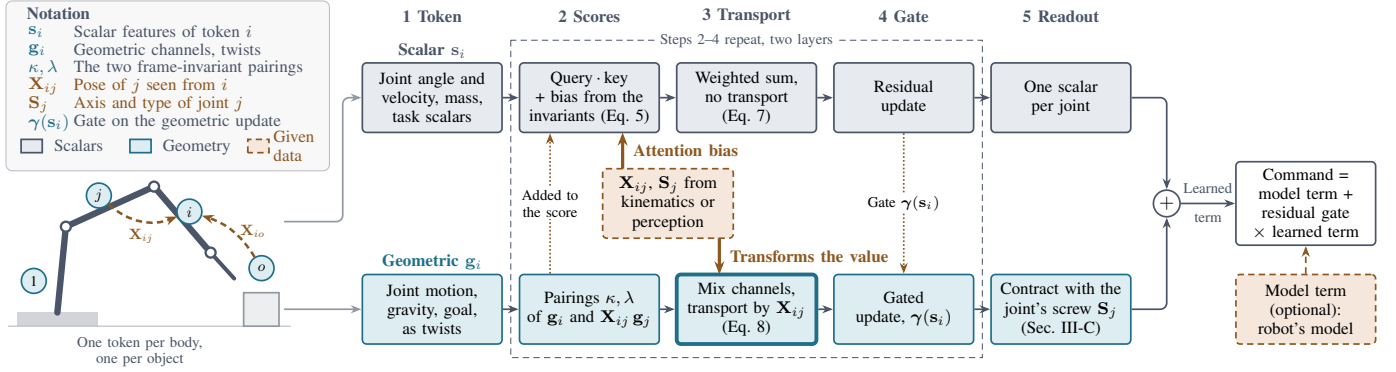
\begin{figure*}[!t]
\centering
\resizebox{\textwidth}{!}{%
\begin{tikzpicture}[
  x=1cm, y=1cm, font=\small,
  lk/.style={line width=3pt, draw=lane-s, line cap=round, line join=round},
  jt/.style={circle, draw=lane-s, fill=white, line width=1.1pt, inner sep=0pt,
             minimum size=6pt},
  num/.style={circle, draw=lane-g, fill=fill-g, thick, inner sep=0.5pt,
              minimum size=15pt, font=\small},
  onum/.style={num, draw=lane-p, fill=fill-p},
  pose/.style={-{Stealth[length=2mm]}, draw=lane-p, densely dashed,
               line width=1.2pt},
  feed/.style={-{Stealth[length=1.8mm]}, draw=lane-s!60, thick},
  sym/.style={font=\footnotesize, text=lane-g, anchor=west},
  dsc/.style={font=\footnotesize, text=lane-s, anchor=west},
  symp/.style={sym, text=lane-p},
  dscp/.style={dsc, text=lane-p},
  op/.style={draw, thick, rounded corners=2pt, align=center, font=\footnotesize,
             minimum height=1.15cm, text width=2.1cm, inner sep=3pt},
  ops/.style={op, draw=lane-s, fill=fill-s},
  opg/.style={op, draw=lane-g, fill=fill-g},
  opp/.style={op, draw=lane-p, fill=fill-p, densely dashed},
  ours/.style={op, draw=lane-g, fill=fill-g, line width=1.8pt},
  head/.style={font=\footnotesize\bfseries, text=lane-s, align=center},
  lane/.style={font=\footnotesize\bfseries, align=center},
  flow/.style={-{Stealth[length=1.8mm]}, thick, draw=lane-s},
  cross/.style={-{Stealth[length=1.8mm]}, thick, draw=lane-p, densely dotted}]

\begin{scope}[shift={(-6.2,-3.85)}, scale=0.78, every node/.append style={transform shape}]
  \fill[lane-s!25] (-0.85,0.09) rectangle (0.85,0.38);
  \draw[lane-s!45] (-1.05,0.09) -- (4.95,0.09);
  \draw[lk] (0,0.38) -- (0.18,2.20) -- (2.05,3.05) -- (3.25,1.65);
  \draw[lk, line width=1.9pt] (3.25,1.65) -- (3.75,1.10);
  \node[jt] at (0.18,2.20) {}; \node[jt] at (2.05,3.05) {};
  \node[jt] at (3.25,1.65) {};
  \node[num] at (-0.50,1.05) {1};
  \node[num] (bj) at (0.88,2.86) {$j$};
  \node[num] (bi) at (2.80,2.52) {$i$};
  \draw[thick, draw=lane-s!60, fill=lane-s!10] (3.95,0.09) rectangle (4.70,0.80);
  \node[num] (bo) at (4.33,1.32) {$o$};
  \draw[pose] (bj) to[bend right=32]
      node[midway, below=1pt, text=lane-p] (xij) {$\X_{ij}$} (bi);
  \draw[pose] (bo) to[bend right=20]
      node[midway, right=1pt, text=lane-p] {$\X_{io}$} (bi);
  \node[font=\small, text=lane-s, align=center] at (1.75,-0.45)
    {One token per body,\\ one per object};
\end{scope}

\begin{scope}[shift={(-6.85,-0.85)}]
  \draw[draw=lane-s!40, fill=lane-s!4, rounded corners=3pt]
        (-0.20,-0.36) rectangle (5.15,2.45);
  \node[font=\footnotesize\bfseries, text=lane-s, anchor=west] at (0.02,2.24)
    {Notation};
  \node[sym]  at (0.05,1.94) {$\mathbf{s}_i$};
  \node[dsc]  at (0.80,1.94) {Scalar features of token $i$};
  \node[sym]  at (0.05,1.64) {$\mathbf{g}_i$};
  \node[dsc]  at (0.80,1.64) {Geometric channels, twists};
  \node[sym]  at (0.05,1.34) {$\kappa, \lambda$};
  \node[dsc]  at (0.80,1.34) {The two frame-invariant pairings};
  \node[symp] at (0.05,1.04) {$\X_{ij}$};
  \node[dscp] at (0.80,1.04) {Pose of $j$ seen from $i$};
  \node[symp] at (0.05,0.74) {$\mathbf{S}_j$};
  \node[dscp] at (0.80,0.74) {Axis and type of joint $j$};
  \node[sym]  at (0.05,0.44) {$\boldsymbol{\gamma}(\mathbf{s}_i)$};
  \node[dsc]  at (0.80,0.44) {Gate on the geometric update};
  \draw[draw=lane-s, thick, fill=fill-s] (0.05,-0.06) rectangle (0.40,0.16);
  \node[dsc, anchor=mid west] at (0.47,0.05) {Scalars};
  \draw[draw=lane-g, thick, fill=fill-g] (1.85,-0.06) rectangle (2.20,0.16);
  \node[dsc, anchor=mid west] at (2.27,0.05) {Geometry};
  \draw[draw=lane-p, thick, densely dashed, fill=fill-p] (3.80,-0.06) rectangle (4.15,0.16);
  \node[dscp, anchor=west, align=left, inner sep=0pt] at (4.22,0.05) {Given\\ data};
\end{scope}

\foreach \i/\t in {0/{1 Token},1/{2 Scores},2/{3 Transport},3/{4 Gate},
                   4/{5 Readout}}
  \node[head] at (2.6*\i, 1.35) {\t};
\node[lane, text=lane-s] at (0, 0.78)  {Scalar $\mathbf{s}_i$};
\node[lane, text=lane-g] at (0,-2.72)  {Geometric $\mathbf{g}_i$};

\node[ops] (s1) at (0,0)    {Joint angle and\\ velocity, mass,\\ task scalars};
\node[ops] (s2) at (2.6,0)  {Query$\,\cdot\,$key\\ + bias from the\\ invariants (Eq.~\ref{eq:scores})};
\node[ops] (s3) at (5.2,0)  {Weighted sum,\\ no transport\\ (Eq.~\ref{eq:scalar})};
\node[ops] (s4) at (7.8,0)  {Residual\\ update};
\node[ops] (s5) at (10.4,0) {One scalar\\ per joint};

\node[opp, text width=1.95cm] (pr) at (3.9,-1.75)
  {$\X_{ij}$, $\mathbf{S}_j$ from\\ kinematics or\\ perception};

\node[opg]  (g1) at (0,-3.5)    {Joint motion,\\ gravity, goal,\\ as twists};
\node[opg]  (g2) at (2.6,-3.5)  {Pairings $\kappa, \lambda$\\ of $\mathbf{g}_i$ and $\X_{ij}\,\mathbf{g}_j$};
\node[ours] (g3) at (5.2,-3.5)  {Mix channels,\\ transport by $\X_{ij}$\\
                                 (Eq.~\ref{eq:update})};
\node[opg]  (g4) at (7.8,-3.5)  {Gated\\ update, $\boldsymbol{\gamma}(\mathbf{s}_i)$};
\node[opg]  (g5) at (10.4,-3.5) {Contract with the\\ joint's screw $\mathbf{S}_j$\\ (Sec.~\ref{sec:readouts})};

\node[opp] (p3) at (14.45,-3.5) {Model term\\ (optional):\\ robot's model};
\node[op, draw=lane-s, fill=white] (tau) at (14.45,-1.75)
  {Command =\\ model term +\\ residual gate\\ $\times$ learned term};
\node[circle, draw=lane-s, thick, fill=white, inner sep=0pt, minimum size=0.44cm,
      font=\small] (plus) at (12.15,-1.75) {$+$};

\foreach \a/\b in {s1/s2, s2/s3, s3/s4, s4/s5, g1/g2, g2/g3, g3/g4, g4/g5}
  \draw[flow] (\a) -- (\b);
\draw[flow, rounded corners=3pt] (s5.east) -- (12.15,0) -- (plus.north);
\draw[flow, rounded corners=3pt] (g5.east) -- (12.15,-3.5) -- (plus.south);
\draw[flow] (plus.east) -- node[midway, above=1pt, font=\scriptsize, text=lane-s] {Learned}
                           node[midway, below=1pt, font=\scriptsize, text=lane-s] {term}
            (tau.west);
\draw[flow, draw=lane-p, densely dashed] (p3.north) -- (tau.south);

\draw[-{Stealth[length=2.2mm]}, draw=lane-p, line width=1.5pt]
  (3.15,-1.175) -- (3.15,-0.575)
  node[midway, right=1pt, font=\footnotesize\bfseries, text=lane-p]
  {Attention bias};
\draw[-{Stealth[length=2.2mm]}, draw=lane-p, line width=1.5pt]
  (4.75,-2.325) -- (4.75,-2.925)
  node[midway, right=1pt, font=\footnotesize\bfseries, text=lane-p]
  {Transforms the value};

\draw[cross] ([xshift=-0.65cm]g2.north) --
  node[midway, font=\scriptsize, fill=white, inner sep=1pt, align=center]
  {Added to\\ the score}
  ([xshift=-0.65cm]s2.south);
\draw[cross] (s4.south) -- node[font=\scriptsize, fill=white, inner sep=1pt]
  {Gate $\boldsymbol{\gamma}(\mathbf{s}_i)$} (g4.north);
\draw[lane-s, densely dashed] (1.3,0.95) rectangle (9.1,-4.3);
\node[font=\scriptsize, text=lane-s, fill=white, inner sep=1pt] at (5.2,0.95)
  {Steps 2--4 repeat, two layers};

\draw[feed, rounded corners=3pt] (-2.45,-2.05) -- (-1.48,-2.05) -- (-1.48,0) -- (s1.west);
\draw[feed] (-2.45,-3.5) -- (g1.west);
\end{tikzpicture}}
\vspace{-0.9\baselineskip}
\caption{\textbf{The Screw Attention layer}, read from left to right. (1) Every body, robot link or object, is a token with a scalar lane $\mathbf{s}_i$ and a geometric lane $\mathbf{g}_i$ of twists. The \textbf{pair term} (orange), the relative pose $\X_{ij}$ and, for robot pairs, the joint screw $\mathbf{S}_j$, comes from the robot model or from perception and is never learned. (2) One set of attention weights serves both lanes and sees the pair term only through invariant quantities. (3) Geometric messages are transformed by $\X_{ij}$ into the receiver's frame; $\mathbf{W}_g$ mixes channels only. (4) The gate $\boldsymbol{\gamma}(\mathbf{s}_i)$ undoes the softmax normalisation. (5) The readout pairs the channels with each joint's screw to produce a joint torque; an optional model term enters through a gate (Section~\ref{sec:readouts}). Head groups and wrench channels are not drawn.}
\label{fig:layer}
\end{figure*}

\textbf{Kinematics-structured policies.} A third strand brings the robot's kinematics inside the network, beyond its connectivity. XMoP~\cite{xmop2024} tokenises the
links as poses and predicts link-wise pose transformations to transfer across manipulators. The Rodrigues Network~\cite{rodrinet2025} makes forward kinematics a learnable operator as
a prior for action learning, and ABD-NET~\cite{abdnet2026} mirrors the articulated-body algorithm in a graph network with learnable inertia parameters. These methods carry the same
quantities that our layer uses, but XMoP and the Rodrigues Network process them with learned matrices applied directly to geometric coordinates, and ABD-NET replaces the edges and the inertias by learned ones. These operations do not in general commute with a change of frame. The output can then depend on how the reference frames are attached to the links, a modelling convention rather than a physical property. XMoP compensates for this with frame-randomising data augmentation and reports losing $68.1\%$ of its success rate on unseen embodiments without it. Screw Attention, instead, keeps the transforms, screws and inertias as the model's own quantities and the geometry inside operations that commute with a change of frame. Messages are transported by the relative pose between bodies instead of being mapped from their coordinates. This also makes the velocity recursion an exact special case of the layer (Section~\ref{sec:representability}) rather than a function it may approximate. 

\textbf{Physics outside the architecture.} A final strand introduces physical knowledge without constraining the network itself. The policy of~\cite{embodyequi2025} achieves
equivariance to changes in the definitions of the end-effector and camera frames through an analytic decoder. This is possible because its inputs, images and language, do not change
when these frames are redefined. In our setting, however, altering the per-link frames changes every quantity the policy reads, so the invariance must be a property of the network rather than the decoder.

Other approaches keep physics in the objective or in a separate model. PIPER~\cite{piper2026} regularises the policy objective with analytic dynamics, and KING~\cite{king2026} represents the robot as a kinematic graph but estimates its motion instead of predicting commands. By contrast, residual policy learning~\cite{silver2018residual,johannink2019residual}
keeps an analytic controller at run time and learns a correction added to its output with an unstructured network. These approaches constrain the objective, the
prediction target or the command, and are actually compatible with the constraint on the function class proposed in this paper.

\section{Method}
\label{sec:method}

Fig.~\ref{fig:layer} shows our architecture, read from left to right. Each body of the robot and each object in the scene is defined as a token. The attention layer updates each token in two lanes, one for scalar quantities and one for geometric quantities, which share the same attention weights.
A readout then turns the tokens into the command, optionally added to an analytic controller.

\subsection{Tokens and the pair term}
\label{sec:tokens}

We describe rigid-body quantities with the spatial vector algebra~\cite{featherstone2008}. In this algebra, the velocity of a body is a six-dimensional twist $\mathbf{m} = [\boldsymbol{\omega}; \mathbf{v}]$, which combines
its angular and linear velocity,  while the load on a body is a wrench $\mathbf{f} = [\mathbf{n}; \boldsymbol{\phi}]$, which combines a moment and a force. Twists and wrenches are both screws. Each is expressed in the coordinate frame of a particular body, and two such quantities can be combined only when they are written in the same frame. A twist expressed in the frame of body $j$ is transformed into the frame of body $i$ by the $6\times6$ spatial transform $\X_{ij}$, and a wrench by  the dual transform $\X_{ij}^{-\top}$. 
Assuming that each body of the robot is driven by one joint, we index the joint by the body it drives. A body is moved by its own joint but also by every joint between it and the base. The joint of body $j$ is described by its motion subspace $\mathbf{S}_j$, expressed in the frame
of body $j$. For a revolute joint with unit axis $\mathbf{a}$ passing through the point
$\mathbf{p}$, $\mathbf{S}_j = [\mathbf{a}; \mathbf{p}\times\mathbf{a}]$. For a prismatic joint along $\mathbf{a}$,
$\mathbf{S}_j = [\mathbf{0}; \mathbf{a}]$. Thus $\mathbf{S}_j$ encodes both the type and the axis of the joint. With $q_j$ the position of the joint (an angle for a revolute joint and a displacement for a
prismatic one), the product $\mathbf{S}_j \dot q_j$ is the twist that the joint adds to the body it
drives.

\textbf{Tokens.} Each body of the robot and each object in the scene is represented by a
token. A token receives two kinds of input. The first is a vector $\mathbf{x}_i$ of
frame-independent quantities, such as the joint position and velocity, the mass, the number of joints between the body and the base, and the gripper state. The second is a set of input twists expressed in the frame of token $i$,
such as the joint's twist $\mathbf{S}_i \dot q_i$, the direction of gravity or the relation to a
goal. A direction or a displacement enters as a twist with a zero angular part, for example
gravity as $[\mathbf{0}; \mathbf{d}]$ with $\mathbf{d}$ the downward direction. A rotation enters as a
twist with a zero linear part. The input twists are stacked as the rows of
$\mathbf{M}_i \in \mathbb{R}^{n_{\mathrm{in}}\times6}$. Each kind of input
is embedded separately,
\begin{equation}
\mathbf{s}_i = \mathbf{W}_{s}\, \mathbf{x}_i, \qquad
\mathbf{g}_i = \mathbf{W}_{\mathrm{in}}\, \mathbf{M}_i,
\label{eq:embed}
\end{equation}
with both matrices $\mathbf{W}_{s}$ and $\mathbf{W}_{\mathrm{in}}$ learned. The scalar features $\mathbf{s}_i \in \mathbb{R}^{d}$ are the state of
the scalar lane. The geometric channels $\mathbf{g}_i \in \mathbb{R}^{k\times6}$ are $k$ twists
expressed in the frame of token $i$. Each channel is a weighted sum of the input twists in $\mathbf{M}_i$. The width $d$ and the number of channels
$k$ are hyperparameters (Section~\ref{sec:config}).

Optionally, a token also holds $k_f$ wrench
channels $\mathbf{g}^f_i \in \mathbb{R}^{k_f\times6}$ and the spatial inertia
$\mathbf{I}_i \in \mathbb{R}^{6\times6}$. The wrench channels are embedded in the same way as the geometric channels. The measured wrenches, stacked as the rows of $\mathbf{F}_i$ are mapped by:
$\mathbf{g}^f_i = \mathbf{W}^f_{\mathrm{in}}\, \mathbf{F}_i$. If no wrench is measured, $\mathbf{g}^f_i = \mathbf{0}$. The spatial inertia encodes the mass, centre of mass and
rotational inertia of the body. For a robot body, it is supplied by the robot model.
For an object, it is set to zero.

\textbf{The pair term.} The pair term is defined for every ordered pair of tokens $(i, j)$, including $i = j$, with $i$ the receiver and $j$ the sender. Each pair of tokens $(i,j)$ is associated with the spatial transform $\X_{ij}$. This transform encodes the pose of body $j$ relative to body $i$. It transforms a
twist from the frame of $j$ into the frame of $i$. When $j$ is a body of the
robot, the pair is also associated with the motion subspace $\mathbf{S}_j$ of its joint. For the
robot, $\X_{ij}$ is computed from the joint positions by forward kinematics at every time step, while
$\mathbf{S}_j$ remains constant in the frame of body $j$.
For an object in the scene, $\X_{ij}$ is computed from its pose, which can generally be estimated by a perception system. A pair whose sender $j$ is an object carries only $\X_{ij}$ while a pair whose receiver $i$ is an object is used only when $j = i$. The pair term is an input to the network and is not learned.

\subsection{The attention layer}
\label{sec:layer}

\textbf{Pairings.} In standard transformers, the attention weights depend only on the token features. In our approach, we also want these attention weights to depend on the geometric relation between what two tokens carry, such as their current velocities, a joint axis or the goal. To make this relation frame-independent, we use the two frame-invariant bilinear forms on twists: the Killing form $\kappa$ and the Klein form $\lambda$ (the reciprocal product of screw theory)~\cite{selig2005geometric,minguzzi2013screw}, which we call pairings. For
two twists $\mathbf{m} = [\boldsymbol{\omega}; \mathbf{v}]$ and $\mathbf{m}' = [\boldsymbol{\omega}'; \mathbf{v}']$
expressed in the same frame, these pairings are
\begin{equation}
\kappa(\mathbf{m}, \mathbf{m}') = \boldsymbol{\omega} \cdot \boldsymbol{\omega}', \qquad
\lambda(\mathbf{m}, \mathbf{m}') = \boldsymbol{\omega} \cdot \mathbf{v}' + \mathbf{v} \cdot \boldsymbol{\omega}'.
\label{eq:pairings}
\end{equation}
The following two examples show what the pairings measure in specific cases. For a joint axis $\mathbf{S} = [\mathbf{a}; \mathbf{p}\times\mathbf{a}]$ and a twist $\mathbf{m}$,
\begin{equation}
\kappa(\mathbf{S}, \mathbf{m}) = \mathbf{a}\cdot\boldsymbol{\omega}, \qquad
\lambda(\mathbf{S}, \mathbf{m}) = \mathbf{a}\cdot(\mathbf{v} + \boldsymbol{\omega}\times\mathbf{p}).
\label{eq:pair-axis}
\end{equation}
These are the rate at which $\mathbf{m}$ rotates about
the axis, and the rate at which $\mathbf{m}$ slides along the axis.
For two rotations about axes $\mathbf{a}$ and $\mathbf{a}'$ through points $\mathbf{p}$ and $\mathbf{p}'$, 
\begin{equation}
\kappa = \mathbf{a}\cdot\mathbf{a}' = \cos\theta, \qquad
\lambda = (\mathbf{p}-\mathbf{p}')\cdot(\mathbf{a}\times\mathbf{a}') = \pm\,\delta\sin\theta,
\label{eq:pair-axes}
\end{equation}
where $\theta$ is the angle between the axes and $\delta$ the distance between them. Every frame-independent symmetric bilinear form on twists is a weighted sum of the pairings $\kappa$ and $\lambda$~\cite{selig2005geometric}, so no other pairing is needed. Note that the dot product of a standard transformer is not used here since, applied to twists, it is frame-dependent. Moving the origin of the frame by $\mathbf{r}$ changes the linear part of each twist from $\mathbf{v}$ to $\mathbf{v} + \boldsymbol{\omega}\times\mathbf{r}$, and therefore changes $\mathbf{m}^{\top}\mathbf{m}'$. By contrast, the pairings $\kappa$ and $\lambda$ are preserved under any change of frame.

Accordingly, in the layer, a twist $\mathbf{m}_j$ of token $j$ is first transformed into the frame of
token $i$ as $\X_{ij}\mathbf{m}_j$. In the frame of token $i$, the pairings
$\kappa(\mathbf{m}_i, \X_{ij}\mathbf{m}_j)$ and $\lambda(\mathbf{m}_i, \X_{ij}\mathbf{m}_j)$ with a twist
$\mathbf{m}_i$ of token $i$ give two frame-independent values that relate the two tokens $i$ and $j$. These values are then used in the attention weights. 

\textbf{Masks.} The $H$ attention heads of each layer are split into two groups. An \emph{outward} head lets
a robot token attend to itself and to the tokens of its ancestors in the kinematic tree. An
\emph{inward} head lets a robot token attend to itself and to the tokens of its descendants. Body $j$ is an ancestor of body $i$ if the joint of $j$ moves $i$, and body $i$ is then a
descendant of $j$. These two attention groups mirror the two sweeps of the recursive
Newton--Euler algorithm~\cite{luh1980online,featherstone2008}, which computes the velocities of the bodies from the base outward and their wrenches from the outermost bodies back toward the base. Additionally, every robot token attends to every scene token, in both groups, while a scene token attends only to itself. We write $\mathcal{M}(i)$ for the set of
tokens that token $i$ may attend to in a given head. Pairs outside $\mathcal{M}(i)$ receive no attention weight; the token $i$ itself is in $\mathcal{M}(i)$ in every head.

\textbf{Attention weights.} For clarity, we omit the head index in what follows. Let
$\mathbf{g}_i^c$ denote channel $c$ of the geometric channels $\mathbf{g}_i$. The attention weights $\alpha_{ij}$ are then defined as those of a standard transformer~\cite{vaswani2017attention} with an additional term per pair (step~2 in Fig.~\ref{fig:layer}),
\begin{align}\alpha_{ij} &= \operatorname{softmax}_{j \in \mathcal{M}(i)} \big( e_{ij} \big), \nonumber \\
e_{ij} &= \tfrac{1}{\sqrt{d_h}}\, (\mathbf{W}_q \mathbf{s}_i)^{\!\top} (\mathbf{W}_k \mathbf{s}_j) + \mathbf{w}_e^{\top} \boldsymbol{\varphi}_{ij}
\label{eq:scores} \\
&\quad + \textstyle\sum_{c} a_c\, \kappa\big(\mathbf{g}_i^c,\, \X_{ij}\, \mathbf{g}_j^c\big)
  + \sum_{c} b_c\, \lambda\big(\mathbf{g}_i^c,\, \X_{ij}\, \mathbf{g}_j^c\big), \nonumber
\end{align}
where $d_h$ is the width of a head. The first term is the query--key product of a standard
transformer, with learned matrices $\mathbf{W}_q$ and $\mathbf{W}_k$. The second term adds the pair
features $\boldsymbol{\varphi}_{ij} \in \mathbb{R}^{14}$, weighted by the learned vector $\mathbf{w}_e$. These
features are given, not learned. They describe the relative placement of the two frames,
the relation of their joint axes (Eq.~\ref{eq:pair-axes}) and the structure of the kinematic tree. The last two terms
pair channel $c$ of token $i$ with channel $c$ of token $j$, transformed into the frame of $i$,
where the weights $a_c$ and $b_c$ are learned. When wrench channels are present, the score receives three further terms,
\begin{equation}
\begin{aligned}
e_{ij} \mathrel{+}= {} & \textstyle\sum_{c} a^f_c\, \kappa_f\big(\mathbf{g}^{f,c}_i,\, \X_{ij}^{-\top}\mathbf{g}^{f,c}_j\big) \\
 & + \textstyle\sum_{c} b^f_c\, \lambda\big(\mathbf{g}^{f,c}_i,\, \X_{ij}^{-\top}\mathbf{g}^{f,c}_j\big) \\
 & + \textstyle\sum_{c,c'} d_{cc'}\, (\mathbf{g}^{c}_i)^{\top} \X_{ij}^{-\top}\mathbf{g}^{f,c'}_j,
\end{aligned}
\label{eq:scores-wrench}
\end{equation}
with learned weights $a^f_c$, $b^f_c$ and $d_{cc'}$. The wrench channels of token $j$ are transported into the frame of $i$ by the dual transform, as in Eq.~\ref{eq:wrench}. The first
two terms pair each wrench channel of $i$ with the same channel of $j$: $\lambda$ applies to wrenches unchanged, and $\kappa_f$ is $\kappa$ taken on the force parts, $\boldsymbol{\phi}\cdot\boldsymbol{\phi}'$, because a change of frame only rotates the force part of a wrench. The third term is the power
$\mathbf{m}^{\top}\mathbf{f}$ between each geometric channel of $i$ and each wrench channel of $j$. The power is the rate at which a wrench does work on a motion, and it is the frame-independent way to pair a twist with a wrench, since $(\X_{ij}\mathbf{m})^{\top}(\X_{ij}^{-\top}\mathbf{f}) = \mathbf{m}^{\top}\mathbf{f}$.
Appendix~\ref{app:features} details the pair features $\boldsymbol{\varphi}_{ij}$.

\textbf{Updates.} The scalar features are updated by standard attention, with learned value
and output matrices $\mathbf{W}_v$ and $\mathbf{W}_o$, a residual connection and no feed-forward block,
\begin{equation}
\mathbf{s}_i \leftarrow \mathbf{s}_i + \mathbf{W}_o \Big[ \textstyle\sum_j \alpha_{ij}\, \mathbf{W}_v \mathbf{s}_j \Big].
\label{eq:scalar}
\end{equation}
The geometric channels are updated with the same attention weights, but by transported
messages (step~3 in Fig.~\ref{fig:layer}),
\begin{equation}
\mathbf{g}_i \leftarrow \mathbf{g}_i + \boldsymbol{\gamma}(\mathbf{s}_i) \odot \mathbf{W}_{go} \Big[ \textstyle\sum_j \alpha_{ij}\,
  \X_{ij}\, (\mathbf{W}_g\, \mathbf{g}_j) \Big].
\label{eq:update}
\end{equation}
The message from token $j$ mixes the channels of $\mathbf{g}_j$ with the learned matrix $\mathbf{W}_g$,
transforms them into the frame of $i$ with $\X_{ij}$, and is weighted by $\alpha_{ij}$. The learned
matrix $\mathbf{W}_{go}$ mixes the channels of the sum, and the gate
$\boldsymbol{\gamma}(\mathbf{s}_i) = \mathbf{W}_{\gamma} \mathbf{s}_i + \mathbf{b}_{\gamma} \in \mathbb{R}^{k}$ scales each geometric channel through the product $\odot$.
The learned matrices act on the channel index, and only the given transform $\X_{ij}$ acts on
the six components. This
makes the update independent of the choice of frames (Section~\ref{sec:invariance}).
As in a standard transformer, each head has its own weights, and the outputs of the heads are
concatenated. The network stacks $L$ such layers.

\textbf{The gate.} The gate $\boldsymbol{\gamma}$ (step~4 in Fig.~\ref{fig:layer}) lets one layer
compute the velocity of every body from the joint velocities. This computation is the outward
sweep of the recursive Newton--Euler algorithm. With $\mathcal{A}(i)$ the set containing body
$i$ and its ancestors, by the velocity recursion, the velocity twist of body $i$ is
\begin{equation}
\mathbf{m}_i = \textstyle\sum_{j \in \mathcal{A}(i)} \X_{ij}\, \mathbf{S}_j\, \dot q_j .
\label{eq:velocity}
\end{equation}
The update of Eq.~\ref{eq:update} computes a sum of the same kind. In an outward head, the robot
tokens that token $i$ attends to are those in $\mathcal{A}(i)$, the same set as in Eq.~\ref{eq:velocity}. The message from token $j$ can carry its joint twist
$\mathbf{S}_j \dot q_j$, since that twist is one of the inputs of token $j$ (Eq.~\ref{eq:embed}) and $\mathbf{W}_{g}$ can select it.  Finally, $\X_{ij}$ transforms the message into the frame of $i$, as in Eq.~\ref{eq:velocity}. The two sums differ only
in their weights. Eq.~\ref{eq:velocity} weights each term by one, whereas the softmax $\alpha_{ij}$ of a
standard transformer makes the attention weights sum to one, so that each term receives
$1/|\mathcal{A}(i)|$. We keep the softmax and add the gate $\boldsymbol{\gamma}$, which scales the channel
by $|\mathcal{A}(i)|$ and so removes this difference. The gate can produce this factor because it
is not bounded to $[0,1]$ and because the scalar features include $|\mathcal{A}(i)|$, the number of joints between body $i$ and the base. Since $\boldsymbol{\gamma}$ depends only on the frame-independent
scalar features, the scaled geometric channel transforms under a change of frame exactly as the unscaled
one. With the gate, one layer reproduces Eq.~\ref{eq:velocity} exactly when the tokens are the bodies of the robot
(Section~\ref{sec:representability}).

\textbf{Wrench channels.} When present, the wrench channels are updated in two steps. In the first step,
before the attention weights are computed, each token adds the wrenches induced by its spatial inertia:
\begin{equation}
\mathbf{g}^{f}_i \leftarrow \mathbf{g}^{f}_i + \mathbf{I}_i\,(\mathbf{W}_I\, \mathbf{g}_i) + \mathbf{u}_i \times^{*} \mathbf{I}_i\, \mathbf{u}_i,
\qquad \mathbf{u}_i = \mathbf{W}_u\, \mathbf{g}_i.
\label{eq:inertia}
\end{equation}
The matrices $\mathbf{W}_I$ and $\mathbf{W}_u$ are learned and act on the channel index, like
$\mathbf{W}_g$. Eq.~\ref{eq:inertia} mirrors the Newton--Euler equation of a rigid body, $\mathbf{f} = \mathbf{I}\dot{\mathbf{m}} + \mathbf{m} \times^{*}
\mathbf{I}\mathbf{m}$, which gives the wrench $\mathbf{f}$ that moves a body of spatial inertia
$\mathbf{I}$ with velocity twist $\mathbf{m}$ and acceleration $\dot{\mathbf{m}}$. The operator $\times^{*}$ is the spatial cross product of a twist with a wrench. 

In the second step, the wrench channels exchange messages exactly as in Eq.~\ref{eq:update}, but using
the dual transform:
\begin{equation}
\mathbf{g}^f_i \leftarrow \mathbf{g}^f_i + \boldsymbol{\gamma}^f(\mathbf{s}_i) \odot \mathbf{W}^f_{go} \Big[ \textstyle\sum_j \alpha_{ij}\,
  \X_{ij}^{-\top}\, (\mathbf{W}^f_g\, \mathbf{g}^f_j) \Big].
\label{eq:wrench}
\end{equation}
The attention weights are those of Eq.~\ref{eq:scores}, shared with the geometric channels, and
the gate $\boldsymbol{\gamma}^f$ is defined like $\boldsymbol{\gamma}$. As in the geometric update, the learned matrices act only on the channel index, while $\X_{ij}^{-\top}$, $\mathbf{I}_i$ and the cross product $\times^{*}$ act on
the six components. In particular, twists are turned into wrenches by $\mathbf{I}_i$ alone. Without the
inertia coupling, wrench channels that start at zero would stay at zero, because the message
update is linear in the wrench channels.

\subsection{Readouts and the residual}
\label{sec:readouts}

After the final layer, a readout generates the command sent to the robot. Depending on the task, the command may take the form of one torque $\tau_j$ per joint, or a twist at the end-effector. Three readouts are available: the \emph{motion} and \emph{force} readouts for torques, and the \emph{twist} readout for end-effector twists. Each policy uses the readout that matches its command.
The motion and force readouts both pair the joint screw $\mathbf{S}_j$ with the channels of its
body $j$, but differ in the channels they use. The motion readout uses the geometric channels, through the pairings of Eq.~\ref{eq:pairings}, which takes the form of a Jacobian
transpose. The force readout uses the wrench channels, through the power $\mathbf{S}_j^{\top}\mathbf{f}$
(Eq.~\ref{eq:scores-wrench}), as in the final step of the recursive Newton--Euler algorithm, where the torque at a
joint is the component of the transmitted wrench along the joint axis. In both cases, the same weights apply to every joint. The readout therefore has no parameter
tied to a particular joint, and our network can be applied unchanged to arms with different numbers of degrees of freedom.

\textbf{Motion.} The motion readout emits one torque per joint. For the joint of body $j$,
\begin{equation}
\tau_j = \mathbf{w}^{\top} \mathbf{s}_j + \textstyle\sum_c \big[ \mu_c\, \kappa(\mathbf{S}_j, \mathbf{g}^c_j) + \nu_c\, \lambda(\mathbf{S}_j, \mathbf{g}^c_j) \big] + \beta.
\label{eq:torque}
\end{equation}
The weights $\mathbf{w}$, $\mu_c$ and $\nu_c$ and the bias $\beta$ are learned. For each channel $c$, the pairings $\kappa(\mathbf{S}_j, \mathbf{g}^c_j)$ and
$\lambda(\mathbf{S}_j, \mathbf{g}^c_j)$ are scalars representing the rate at which the twist $\mathbf{g}^c_j$
rotates around the joint axis and the rate at which it slides along it (Eq.~\ref{eq:pair-axis}). The torque is a learned weighted sum of these scalars over the
channels. In Section~\ref{sec:representability}, we show that this readout contains the Jacobian-transpose control law as a special case.

\textbf{Force.} The force readout pairs the joint screw with the wrench channels of its body, so that
\begin{equation}
\tau_j = \textstyle\sum_c w_c\, \mathbf{S}_j^{\top} \mathbf{g}^{f,c}_j + \beta.
\label{eq:force}
\end{equation}
The weights $w_c$ and the bias $\beta$ are learned. Apart from $\beta$, the command is formed only from the wrench channels, which the layer computes using the spatial inertia
$\mathbf{I}_i$ and the dual transforms $\X_{ij}^{-\top}$. The force readout is therefore the only readout through
which $\mathbf{I}_i$ enters the command directly.

\textbf{Twist.} The twist readout emits a twist at the end-effector token $e$, as
\begin{equation}
\boldsymbol{\xi} = \operatorname{softplus}(\mathbf{w}_{\xi}^{\top} \mathbf{s}_e + b_\xi) \textstyle\sum_c w^{\xi}_c\, \mathbf{g}^c_e,
\label{eq:twist}
\end{equation}
together with a gripper command $a_g = \mathbf{w}_{g}^{\top} \mathbf{s}_e + b_g$, with
$\mathbf{w}_{g}$ and $b_g$ learned. The weights
$\mathbf{w}_{\xi}$, $b_\xi$ and $w^{\xi}_c$ are also learned. The angular and linear
parts of $\boldsymbol{\xi}$ are each bounded by their norm: each part $\mathbf{z}$ is replaced by
$\tanh(\lVert \mathbf{z} \rVert)\, \mathbf{z} / \lVert \mathbf{z} \rVert$. The output twist is expressed in the frame of the end-effector token. Outside the network, it is
transformed into the frame expected by the controller.

\textbf{Learning alone or with a controller.} Our layer can be used on its own, or on top of
an analytic controller built from the robot's model. The second option combines their strengths: the controller contributes what the model already provides, such as gravity compensation or a task-space servo, while the layer learns only what the model omits, such as contact. For the joint of body $j$, the command becomes
\begin{equation}
\tau_j = \tau_{\mathrm{model},j} + \sigma(\mathbf{w}_{\sigma}^{\top} \mathbf{s}_j + b_{\sigma})\, \tau_{\mathrm{learned},j},
\label{eq:residual}
\end{equation}
where $\tau_{\mathrm{model},j}$ is the output of the analytic controller,
$\tau_{\mathrm{learned},j}$ is the output of the readout, $\sigma$ is the logistic function, and $\mathbf{w}_{\sigma}$ and $b_{\sigma}$ are learned. We refer to $\sigma(\cdot)$ as the residual gate. The model term enters with weight one, so the controller is part of the architecture, not merely its initialisation, even during training. As a result, closing the residual gate recovers the analytic controller exactly. The learned term is therefore a measurable departure from the model, per joint and per step, while the controller remains available as a fallback. 
This construction follows residual policy learning~\cite{silver2018residual,johannink2019residual}, in which the output of an unstructured network is added to that of an analytic controller (without a gate). Here the
residual is produced by the layer itself, so it is expressed through the robot's geometry and inherits the frame invariance as defined in Section~\ref{sec:invariance}. Because the readout shares its weights across joints, the residual applies to arms with different numbers of joints. The residual gate makes its magnitude a learned per-joint quantity. The residual gate is optional. Without it, the learned term is added directly to the model term, as in residual policy learning. The content of the model term and the use of the gate
are chosen per task (Section~\ref{sec:config}).

\begin{figure}[t]
\centering
\resizebox{\columnwidth}{!}{%
\begin{tikzpicture}[
  x=1cm, y=1cm, font=\small,
  lk/.style={line width=5pt, draw=lane-s, line cap=round, line join=round},
  jt/.style={circle, draw=lane-s, fill=white, line width=1.2pt, inner sep=0pt, minimum size=8pt},
  num/.style={circle, draw=lane-s, fill=fill-s, thick, inner sep=0.5pt, minimum size=13pt, font=\scriptsize},
  outw/.style={-{Stealth[length=2mm]}, line width=1.2pt, draw=lane-g},
  spin/.style={-{Stealth[length=1.5mm]}, line width=1pt, draw=lane-g},
  inw/.style={-{Stealth[length=2mm]}, line width=1.2pt, draw=lane-p},
  err/.style={-{Stealth[length=1.5mm]}, line width=1.1pt, draw=lane-p, densely dashed},
  labg/.style={font=\footnotesize, text=lane-g, inner sep=1pt},
  labp/.style={font=\footnotesize, text=lane-p, inner sep=1pt},
  boxg/.style={draw=lane-g, fill=fill-g, thick, rounded corners=3pt, inner sep=5pt, align=left, font=\footnotesize, text=lane-s},
  boxp/.style={draw=lane-p, fill=fill-p, thick, rounded corners=3pt, inner sep=5pt, align=left, font=\footnotesize, text=lane-s}]

\draw[draw=lane-s!60, fill=lane-s!22, thick] (-0.75,0) rectangle (0.75,0.55);
\node[font=\footnotesize, text=lane-s] at (0,0.275) {base};
\draw[lk] (0,0.55) -- (0.4,1.75) -- (2.5,3.0) -- (4.55,2.15) -- (5.9,1.1);
\node[jt] (j1) at (0.4,1.75) {}; \node[jt] (j2) at (2.5,3.0) {}; \node[jt] (j3) at (4.55,2.15) {};
\coordinate (tip) at (5.9,1.1);
\node[num] (b1) at (1.15,2.65) {1};
\node[num] (b2) at (3.95,2.85) {2};
\node[num] (b3) at (5.55,2.05) {3};

\draw[spin] (j1) ++(-0.28,-0.1) arc (200:440:0.28);
\node[labg, anchor=east] at (0.0,1.95) {$\mathbf{S}_1\dot q_1$};
\draw[spin] (j2) ++(-0.28,0.1) arc (160:400:0.28);
\node[labg, anchor=east] at (2.2,3.1) {$\mathbf{S}_2\dot q_2$};
\draw[spin] (j3) ++(0.28,-0.1) arc (-20:220:0.28);
\node[labg, anchor=south] at (4.55,2.53) {$\mathbf{S}_3\dot q_3$};
\draw[outw] (j1) to[out=105, in=80, looseness=1.05] node[labg, above, pos=0.55] {$\X_{31}$} (b3.north east);
\draw[outw] (j2) to[out=30, in=110, looseness=0.75] node[labg, above, pos=0.5] {$\X_{32}$} (b3.north);
\draw[outw] (j3) -- node[labg, above, pos=0.65] {$\X_{33}$} (b3.west);
\draw[outw, line width=1.7pt] (b3.east) -- ++(0.8,0) node[labg, right] {$\mathbf{m}_3$};

\draw[lane-s, thick] (tip) -- ++(0.5,-0.36);
\draw[lane-s, thick] (tip) -- ++(0.36,0.5);
\draw[lane-p, thick, densely dashed] (tip) -- ++(0.61,0.08);
\draw[lane-p, thick, densely dashed] (tip) -- ++(-0.08,0.61);
\draw[err] ($(tip)+(0.62,-0.45)$) arc (-36:8:0.77);
\node[labp, anchor=west] at ($(tip)+(0.78,-0.16)$) {$\mathbf{e}_\omega$};
\draw[inw] (tip) to[out=215, in=280, looseness=0.8] node[labp, below, pos=0.45] {$\X_{13}$} (j1);
\draw[inw] (tip) to[out=200, in=305, looseness=1.05] node[labp, below, pos=0.4] {$\X_{23}$} (j2);
\draw[inw] (tip) to[out=180, in=270, looseness=1.3] node[labp, left, pos=0.7] {$\X_{33}$} (j3);
\node[labp, anchor=north east] at (0.15,1.25) {$\tau_1$};
\node[labp, anchor=north] at ($(j2)+(-0.08,-0.34)$) {$\tau_2$};
\node[labp, anchor=north east] at (4.3,1.9) {$\tau_3$};

\node[boxg, anchor=north west, text width=3.55cm, minimum height=1.85cm] at (-0.75,-0.2)
  {\textbf{Velocity recursion}\\ one outward head\\[3pt]
   $\mathbf{m}_3 = \sum_{j\in\mathcal{A}(3)} \X_{3j}\mathbf{S}_j\dot q_j$\\[2pt]
   $\alpha_{3j} = \tfrac13$, gate $\times\,3$};
\node[boxp, anchor=north west, text width=4.4cm, minimum height=1.85cm] at (3.55,-0.2)
  {\textbf{Jacobian-transpose law}\\ one inward head, motion readout\\[3pt]
   $\tau_j = \kappa\big(\mathbf{S}_j, \X_{j3}[\mathbf{e}_\omega;\mathbf{0}]\big) = \mathbf{a}_j\!\cdot\mathbf{e}_\omega$\\[2pt]
   $\boldsymbol{\tau} = \mathbf{J}_\omega^{\top}\mathbf{e}_\omega$};
\end{tikzpicture}}
\caption{Representability of the screw attention layer on a three-link arm. Every drawn quantity is given. In \textbf{blue}, the velocity recursion. One outward head carries the joint twists of all ancestors to link 3. The layer expresses this with the gate. In \textbf{orange}, the Jacobian-transpose
law. One inward head can carry the tip error $\mathbf{e}_\omega$ to every ancestor joint. The motion
readout pairs it with each joint axis.}
\label{fig:representability}
\end{figure}
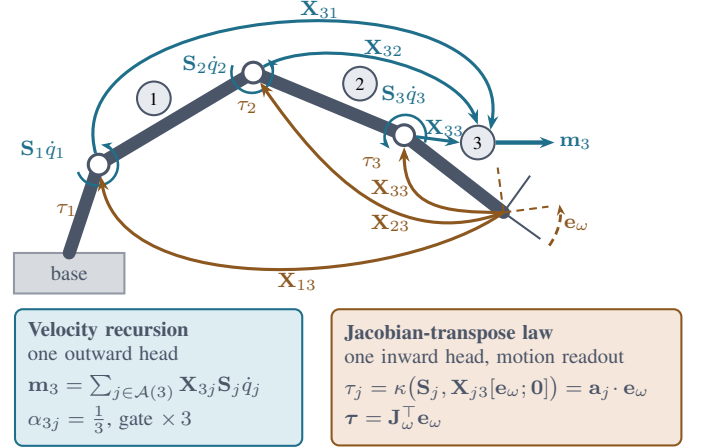

\subsection{Representability}
\label{sec:representability}

Two classical computations from robot mechanics lie within the expressive power of our layer: the
velocity recursion of Eq.~\ref{eq:velocity} and the Jacobian-transpose control
law~\cite{takegaki1981new,lynch2017}. With hand-set weights, our layer reproduces each computation exactly, so a trained policy need not learn them from data. Both constructions share the same structure: a construction sets the few weights required to express a specific computation while leaving every other weight at zero.

For the velocity recursion, a single outward head with uniform attention over the ancestors and a gate that removes the normalisation (Section~\ref{sec:layer}) reproduces Eq.~\ref{eq:velocity} (Fig.~\ref{fig:representability}, blue).

For the Jacobian-transpose law, suppose the task supplies the orientation error $\mathbf{e}_\omega$ of the end-effector as an input twist. An inward head can carry this twist to
body $j$, transformed into the frame of $j$. The $\kappa$ term in Eq.~\ref{eq:torque} (Eq.~\ref{eq:pair-axis}) then equals the $j$-th entry of $\mathbf{J}_\omega^{\top}\mathbf{e}_\omega$, where $\mathbf{J}_\omega$ is the angular Jacobian. A single weight on this channel therefore reproduces the law, on any arm (Fig.~\ref{fig:representability}, orange).
 The same construction with the wrench channels and the force readout gives
$\boldsymbol{\tau} = \mathbf{J}^{\top}\mathbf{F}$ for a wrench $\mathbf{F}$ at the end-effector. The velocity recursion and the force-readout form are validated numerically in Appendix~\ref{sec:verification}.

The velocity recursion and Jacobian-transpose law constructions above use a single layer. Stacking layers extends them to inverse dynamics, which the recursive Newton--Euler algorithm computes in three dependent stages:  first the body velocities, then the accelerations, and finally the wrenches, which are accumulated toward the base and projected onto the joint axes. Because all heads in a layer read the same input, each stage requires its own layer. A stack of three layers, one per stage, is therefore needed to represent the algorithm. The number of layers is determined by these dependent stages and not by the length of the kinematic chain, since the masks give every body direct access to all of its ancestors.  The policies in
this paper use only two layers, and we leave full inverse dynamics with three layers to future work.

\subsection{Frame invariance}
\label{sec:invariance}

We now show the invariance the layer was built for through the following derivation. Let the frame of every token be changed independently, and let $\mathbf{T}_i$ be the $6\times6$ spatial transform associated with the change at token $i$. The inputs to the layer transform as follows:
\begin{equation}
\left\{
\begin{aligned}
\X'_{ij} &= \mathbf{T}_i\, \X_{ij}\, \mathbf{T}_j^{-1}, \\
\mathbf{S}'_j &= \mathbf{T}_j\, \mathbf{S}_j, \\
\mathbf{g}'_i &= \mathbf{T}_i\, \mathbf{g}_i, \\
\mathbf{g}^{f\prime}_i &= \mathbf{T}_i^{-\top}\, \mathbf{g}^f_i, \\
\mathbf{I}'_i &= \mathbf{T}_i^{-\top}\, \mathbf{I}_i\, \mathbf{T}_i^{-1},
\end{aligned}
\right.
\label{eq:frame-change}
\end{equation}
The scalar features remain unchanged because $\mathbf{x}_i$ contains only frame-independent quantities.
Similarly, the pairings $\kappa$ and $\lambda$ and the power $\mathbf{m}^{\top}\mathbf{f}$ remain unchanged under the change of frames.

Every learned matrix acts only on the channel index (Section~\ref{sec:layer}) and therefore commutes with every $\mathbf{T}_i$. The operations on the six components, that is, the transport by $\X_{ij}$, the inertia coupling and the cross products are covariant. For given attention weights, 
the updates of Eqs.~\ref{eq:update}, \ref{eq:inertia} and~\ref{eq:wrench} give
\begin{equation}
\mathbf{g}'_i = \mathbf{T}_i\, \mathbf{g}_i, \qquad
\mathbf{g}^{f\prime}_i = \mathbf{T}_i^{-\top}\, \mathbf{g}^f_i ,
\label{eq:frame-output}
\end{equation}
 that is, the outputs transform exactly as the inputs of Eq.~\ref{eq:frame-change} do, which is the invariance.

 Since the motion and force readouts pair the channels with $\mathbf{S}_j$, the torque command is invariant to the change of frames, up to the three pair features that depend on where the frames are placed on the bodies (Appendix~\ref{app:features}). The twist readout, on the other hand, is invariant to rotations of the end-effector frame only, because its norm bound depends on the point at which the twist is expressed.

 These properties are verified numerically in Appendix~\ref{sec:verification}.

\section{Experiments}
\label{sec:experiments}
\label{sec:pays}

\subsection{Tasks}
\label{sec:tasks}

\begin{figure}[t]
\centering
\begin{tabular}{@{}c@{\hspace{4pt}}c@{}}
\includegraphics[width=0.485\columnwidth]{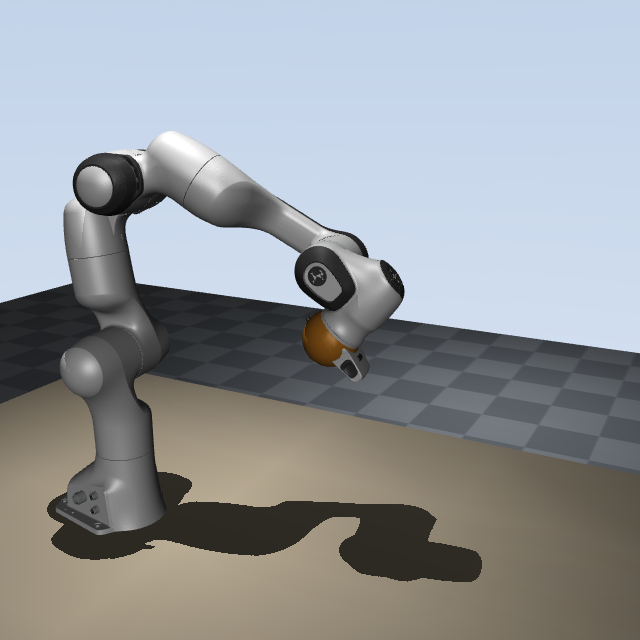} &
\includegraphics[width=0.485\columnwidth]{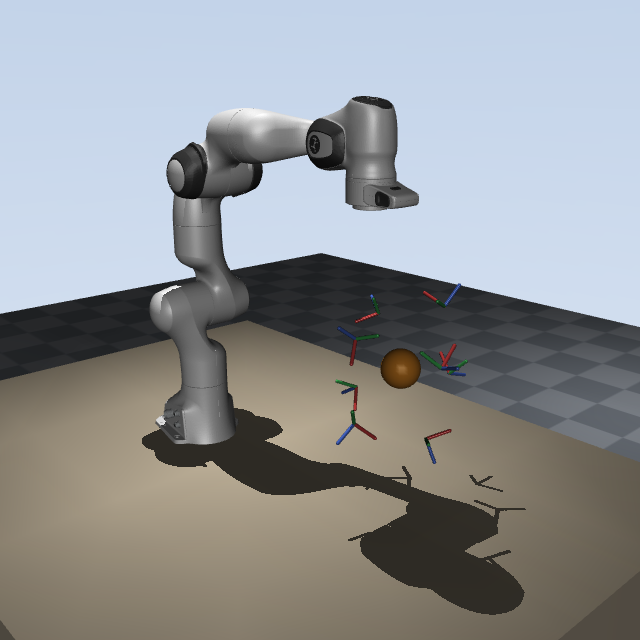} \\[2pt]
\includegraphics[width=0.485\columnwidth]{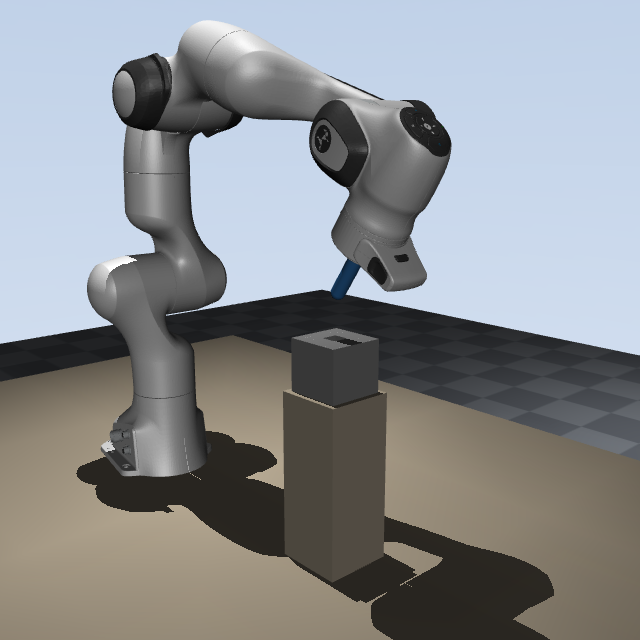} &
\includegraphics[width=0.485\columnwidth]{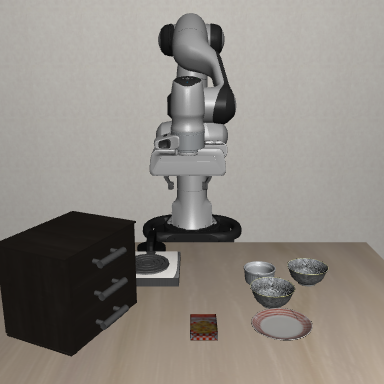} \\
\end{tabular}

\caption{The four tasks, rendered from the environments the results were measured in.
\textbf{Top left}, reorientation: the arm carries a payload (orange) of randomised mass and must bring it to a commanded orientation. \textbf{Bottom left}, peg insertion: the peg (blue)
is welded to the flange and the socket is fixed in space.
\textbf{Top right}, affordance selection: eight candidate approach frames (axis triads) surround an object, an unknown subset is reachable from the current configuration, and the
policy must pick one and go there. \textbf{Bottom right}, one of the ten LIBERO-Spatial tasks at the benchmark's own initial state.
}
\label{fig:tasks}
\end{figure}

We validated Screw Attention on three robot tasks in simulation and one published benchmark (Fig.~\ref{fig:tasks}) described as follows.

\textbf{Reorientation.} This task consisted of reorienting a payload with a fixed-base arm in episodes of 2.0\,s at 50\,Hz. Goals were drawn from one hemisphere of $SO(3)$ in training and the policy evaluated on both. The performance metric was rollout success defined as the angle error $<$0.1 rad held for 20 steps, over 512 fixed episodes.

\textbf{Affordance.} This task consisted of reaching an object along one of the $K=8$ candidate approach frames, distributed over a full sphere around the object. A candidate was considered feasible when the analytic controller, run from the arm's current
configuration, could reach it. In each scene only a subset of the candidates was feasible. The policy was never told which candidates were feasible, and had to select one by
moving toward it without colliding with the object. The performance metric was the fraction of episodes in which a feasible frame was reached without touching the object, over 1024
fixed scenes.

\textbf{Insertion.} This task consisted of inserting a peg welded to the flange of the arm into a socket fixed in space, which involved contact. This setting evaluated the torque residual as both the analytical and learned components were used for the command as specified in Section~\ref{sec:readouts}. The performance metric was insertion success over 768 fixed episodes.

\textbf{LIBERO-Spatial.} Since every comparison on the previous tasks was made against an analytic controller, we also evaluated the approach on a published benchmark.  We used the
ten tasks of LIBERO-Spatial~\cite{libero2023}. In this suite, a named object must be picked up and placed at a destination (within 3\,cm of the destination centre). The source and destination had varying placements. The policy was trained on all ten tasks. The performance metric was the success rate of the placement, evaluated on 500 canonical episodes. Since Screw Attention consumes poses rather than images, the policy uses neither the camera observations nor the language instruction. The poses of all robot links and all objects are read directly from the simulator and form the scene tokens. The policy receives the relation from the gripper to the source and from the source to the destination, in place of the language instruction. This isolates the policy from language understanding and perception, which are outside the scope of this work. Section~\ref{sec:libero} then measures the effect of pose-estimation error by adding noise to these poses.

\subsection{Baselines and ablations}
\label{sec:controls}

To properly evaluate our approach, we designed different comparison networks that either ablate one component of Screw Attention or reimplement the mechanism of a published line of work. Appendix~\ref{app:controls} summarises the baselines and ablations, the mechanism each instantiates, and where each is used.
\begin{itemize}
\item \emph{Shuffled transforms}: the transforms $\X_{ij}$ are permuted across token pairs at every step (on the affordance task, across the robot--scene pairs only), so the correct values are assigned to the wrong pairs.
\item \emph{Scalar only}: the geometric channels are removed and the pair features $\boldsymbol{\varphi}_{ij}$ are kept as score features~\cite{morphformer2026}.
\item \emph{Unconstrained}: the weight matrix $\mathbf{W}_g$ mixes the six spatial components of each geometric channel~\cite{xmop2024,rodrinet2025}.
\item \emph{Full attention}, \emph{adjacency mask} and \emph{random mask}: only the attention
mask changes, to no mask~\cite{gupta2022metamorph}, adjacency in the kinematic tree~\cite{sferrazza2024body}, or a random mask of the same density.
\item \emph{No scene screw} (Affordance): the approach axis is removed from the
candidate tokens. 
\item \emph{Shuffled inertias} (Insertion): the spatial inertias are permuted across bodies.
\item \emph{Flat MLP}: the same observation, flattened, at the matched budget or larger.
\item \emph{Transformer} and \emph{Graph network} (LIBERO-Spatial): a transformer with attention over body tokens~\cite{sferrazza2024body} and a graph network with message passing over the kinematic graph~\cite{abdnet2026}, given the true transforms as edge features, both at the matched
budget.
\item \emph{Analytic controller}: the classical pipeline given the same poses. On the affordance task it selects the candidate nearest the arm.
\end{itemize}
On every robot task, the test fixed before training is that Screw Attention outperforms both shuffled transforms and scalar only.

\subsection{Training and configuration}
\label{sec:config}

For the reorientation and affordance tasks, every policy was trained by behaviour cloning from an analytic demonstrator, then by PPO. The insertion policies were trained by PPO only, since their model term is the demonstrator. Each task was run over several seeds (settings in Appendix~\ref{app:training}). On LIBERO-Spatial every network was fitted to the
benchmark's demonstrations, re-executed with each object independently offset by up to 5\,cm, with the same labels and budget.

All policies used two layers, four heads (two outward, two inward) and $d = 32$. The reorientation policies used $k = 6$ with the motion readout (17{,}948 parameters). The
affordance policy instead used the force readout, with two wrench channels and the inertia coupling (19{,}122 parameters). Its demonstrations were restricted to scenes in which the
demonstrator chose the candidate nearest the arm, and its training used an entropy bonus and a curriculum on the collision penalty. The same strategy was used for all baselines and ablations.

The insertion policies had two wrench channels, the inertia coupling, and the motion readout (18{,}661 parameters). The LIBERO policy
had $k = 16$ with the twist readout (16{,}162 parameters). As in ACT~\cite{zhao2023act}, it predicted a chunk of $N_c = 16$ twists per forward pass and combined them across steps by
temporal ensembling. The twist was converted to a pose change over one control period at the controller's grip site.

Only the affordance policy and the insertion networks had a model term. On the affordance task, the model term held gravity compensation and a velocity brake, without a gate. On the
insertion task, the model term was the task's analytic controller, which servos to the socket pose estimated with a 10\,mm uncertainty and then searches in a spiral. The residual
gate was initialised half open.

\subsection{Comparison with the ablations}
\label{sec:q-pays}

The first test asks whether the geometric content of the pair term, and not added capacity, explains the performance. On each robot task, Screw Attention is compared with the two ablations that remove this content: scalar only drops the geometric channels, and shuffled transforms keeps them but assigns the transforms to the wrong pairs (Table~\ref{tab:results}).

Screw Attention has the highest mean success of the learned networks on all three tasks. On reorientation and affordance, both ablations fall far behind, so the policy depends on transporting the correct relations between bodies. On affordance, Screw Attention also exceeds the analytic controller, which always selects the candidate nearest the arm. The learned policy instead infers which candidate is reachable from the current configuration, and the oracle, which tests every candidate by simulation, bounds what this selection can achieve. On insertion, the ablations stay close to Screw Attention, because the analytic controller in the model term already supplies most of the geometry that the task needs (Section~\ref{sec:doesnot}). The residual still pays on this task: every learned network improves on the analytic controller, and Screw Attention by 17.3 points. The geometric content thus drives the performance wherever the task needs relations between bodies that nothing else supplies.

\begin{table}[t]
\centering
\caption{Comparison with the ablations on the robot tasks: success (\%), mean $\pm$ SD.
Bold: best result in each column. $^\ast$On the affordance task, only the
robot--scene transforms are shuffled (Appendix~\ref{app:controls}).}
\label{tab:results}
\footnotesize
\setlength{\tabcolsep}{3pt}
\begin{tabular}{lcccc}
\toprule
Network & Mechanism & Reorientation & Affordance & Insertion \\
\midrule
Shuffled transforms & --- & 2.7 $\pm$ 1.3 & 6.3 $\pm$ 2.2$^\ast$ & 76.7 $\pm$ 3.2 \\
Scalar only & \cite{morphformer2026} & 2.7 $\pm$ 0.7 & 16.0 $\pm$ 3.6 & 72.9 $\pm$ 2.2 \\
Analytic controller & Classical & \textbf{100.0 $\pm$ 0.0} & 71.5 $\pm$ 0.0 & 62.1 $\pm$ 1.3 \\
\midrule
\textbf{Screw Attention} & --- & 99.2 $\pm$ 0.3 & \textbf{80.2 $\pm$ 1.3}
  & \textbf{79.4 $\pm$ 3.0} \\
\bottomrule
\end{tabular}
\end{table}

\subsection{Ablations of the geometric terms}
\label{sec:where}

\begin{figure}[t]
\centering
\begin{tikzpicture}
\begin{axis}[
  name=aff, scale only axis, width=3.75cm, height=3.4cm,
  xmin=0.45, xmax=6.55, ymin=0, ymax=104,
  xtick={1,2,3,4,5,6},
  xticklabels={SA, SO, NSS, ST, {MLP\\405k}, {MLP\\17k}},
  xticklabel style={align=center, font=\scriptsize},
  ytick={0,20,40,60,80,100}, ylabel={Success (\%)},
  ymajorgrids, grid style={gray!20},
  tick label style={font=\footnotesize}, label style={font=\footnotesize},
  title={(a) Affordance}, title style={font=\footnotesize, yshift=-5pt},
  clip=false,
]
\draw[black!55, densely dashed] (axis cs:0.45,96.2) -- (axis cs:6.55,96.2)
  node[pos=1, anchor=south east, inner sep=1pt, font=\tiny, text=black!75] {Oracle};
\draw[black!55, densely dashed] (axis cs:0.45,71.5) -- (axis cs:6.55,71.5)
  node[pos=1, anchor=south east, inner sep=1pt, font=\tiny, text=black!75] {Analytic};
\draw[black!55, densely dashed] (axis cs:0.45,55.2) -- (axis cs:6.55,55.2)
  node[pos=1, anchor=south east, inner sep=1pt, font=\tiny, text=black!75] {Random};
\addplot[only marks, mark=*, mark size=2.1pt, lane-s,
         error bars/.cd, y dir=both, y explicit] coordinates {
  (2,16.0) +- (0,3.6)
  (3,8.1)  +- (0,4.3)
  (4,6.3)  +- (0,2.2)
  (5,4.7)  +- (0,1.8)
  (6,0.8)  +- (0,0.4)};
\addplot[only marks, mark=*, mark size=2.1pt, lane-g,
         error bars/.cd, y dir=both, y explicit] coordinates {(1,80.2) +- (0,1.3)};
\end{axis}
\begin{axis}[
  at={(aff.south east)}, anchor=south west, xshift=0.85cm,
  scale only axis, width=2.55cm, height=3.4cm,
  xmin=0.45, xmax=4.55, ymin=50, ymax=90,
  xtick={1,2,3,4}, xticklabels={SA, SI, ST, SO},
  xticklabel style={align=center, font=\scriptsize},
  ytick={50,60,70,80,90},
  ymajorgrids, grid style={gray!20},
  tick label style={font=\footnotesize}, label style={font=\footnotesize},
  title={(b) Insertion}, title style={font=\footnotesize, yshift=-5pt},
  clip=false,
]
\draw[black!55, densely dashed] (axis cs:0.45,62.1) -- (axis cs:4.55,62.1)
  node[pos=1, anchor=north east, inner sep=1pt, font=\tiny, text=black!75] {Analytic};
\addplot[only marks, mark=*, mark size=2.1pt, lane-s,
         error bars/.cd, y dir=both, y explicit] coordinates {
  (2,78.1) +- (0,5.5)
  (3,76.7) +- (0,3.2)
  (4,72.9) +- (0,2.2)};
\addplot[only marks, mark=*, mark size=2.1pt, lane-g,
         error bars/.cd, y dir=both, y explicit] coordinates {(1,79.4) +- (0,3.0)};
\end{axis}
\end{tikzpicture}
\caption{Success (\%), mean $\pm$ SD. SA: Screw Attention; NSS: no scene screw; ST:
shuffled transforms; SO: scalar only; SI: shuffled inertias. Dashed lines: random choice (the analytic controller sent to a random candidate), the
analytic controller, and the oracle; in (b), the analytic controller that forms the model
term.}
\label{fig:where}
\end{figure}
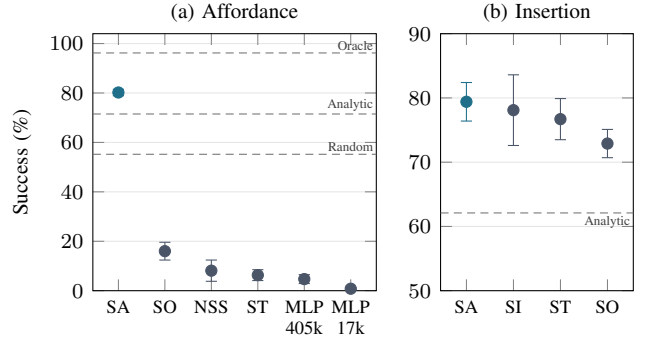

The ablations locate the geometric terms that carry this gain (Fig.~\ref{fig:where}). On affordance, removing the approach axis from the candidate tokens, or shuffling the transforms between the robot and the scene, brings success close to zero. The direction toward a candidate is carried by its screw, transported into the frame of each link, so the policy needs both the screw and the correct transform. Flat MLPs given the same observation also fail, both at the matched budget and at 405k parameters, so the gap is not a matter of capacity. Scalar only does slightly better than these networks, but it reads the action from scalar features and not from forces, so it differs from Screw Attention in more than its geometric channels.

On insertion, the same ablations have smaller effects. Removing the geometric channels costs a few points ($p = 0.014$, Welch's $t$-test), while assigning the transforms to the wrong pairs has no significant effect ($p = 0.27$). Permuting the spatial inertias has no effect either, but this ablation is uninformative: in additional tests, permuting the inertias barely changed the action of the trained policies, so the policies did not use them. The geometric channels carry useful information on this task, but their correctness adds little. Section~\ref{sec:doesnot} discusses why the two tasks differ.

\subsection{Robustness to the frame convention}
\label{sec:constraint}

\begin{figure*}[t]
\centering
{\footnotesize\setlength{\tabcolsep}{3pt}\renewcommand{\arraystretch}{1.35}%
\sbox{\ladderbox}{\begin{tabular}[b]{lcccccc}
\toprule
Network & Mechanism & Params & \hd{Time\\(ms)} & Canonical & \hd{Frame\\change} & \hd{Displaced\\5\,cm} \\
\midrule
Flat MLP & --- & 17{,}572 & 0.02 & 19.9 $\pm$ 9.5 & 0.0 $\pm$ 0.0 & 17.1 $\pm$ 6.4 \\
Transformer & \cite{sferrazza2024body} & 17{,}350 & 0.07 & 75.8 $\pm$ 14.7 & 1.3 $\pm$ 1.3
  & 69.7 $\pm$ 16.8 \\
Graph network & \cite{abdnet2026} & 17{,}038 & 0.18 & 84.7 $\pm$ 5.2 & 2.3 $\pm$ 3.0
  & 75.5 $\pm$ 0.1 \\
Flat MLP & --- & 438{,}384 & 0.05 & 95.2 $\pm$ 1.4 & 0.5 $\pm$ 0.7 & 85.1 $\pm$ 0.1 \\
Analytic controller & Classical & 0 & --- & 81.6 $\pm$ 0.0 & 81.6 $\pm$ 0.0$^\ast$
  & 73.0 $\pm$ 0.0 \\
\midrule
\textbf{Screw Attention} & --- & \textbf{16{,}162} & 0.66 & \textbf{97.3 $\pm$ 0.7}
  & \textbf{97.7 $\pm$ 0.1} & \textbf{86.9 $\pm$ 1.6} \\
\bottomrule
\end{tabular}}%
\begin{tabular}{@{}c@{\hspace{8mm}}c@{}}
\usebox{\ladderbox} &
\includegraphics[height=\dimexpr\ht\ladderbox+\dp\ladderbox\relax]{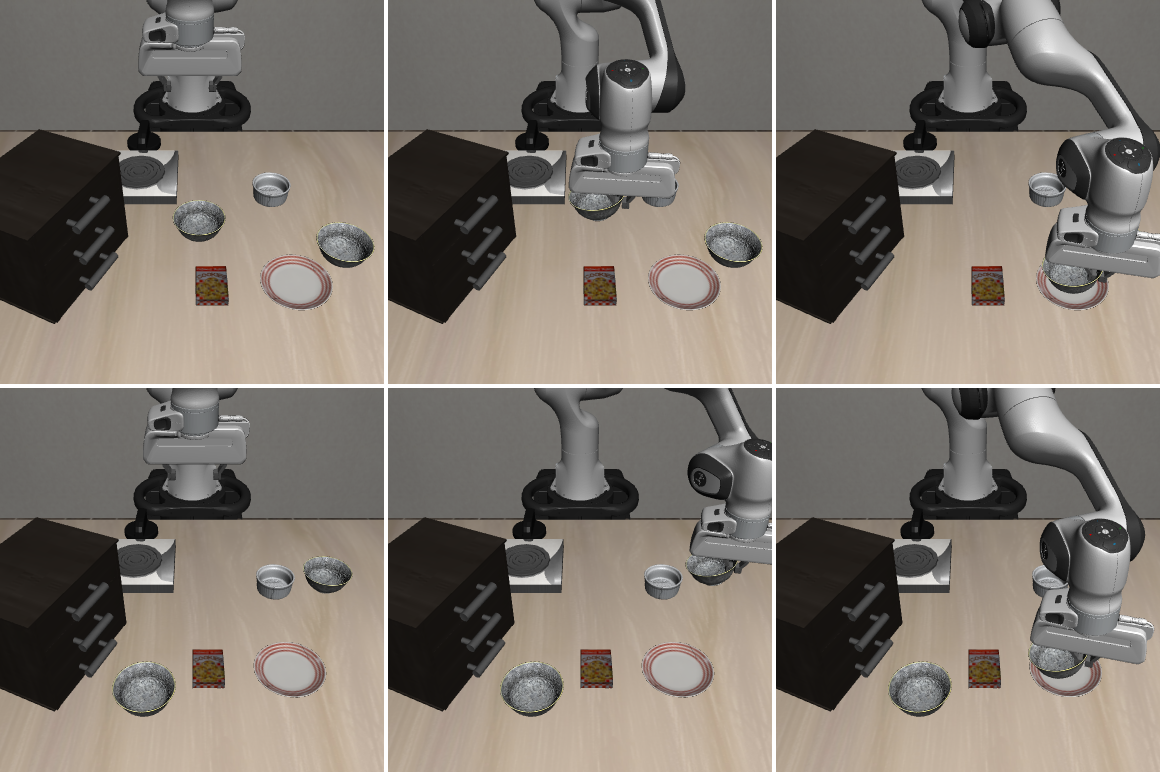} \\[3pt]
(a) & (b)
\end{tabular}}
\caption{LIBERO-Spatial. (a) Success (\%), 500 episodes per run, mean $\pm$ SD. Time: one forward pass on a laptop CPU. Frame change: every token's reference frame changed, policy unchanged. $^\ast$The analytic controller does not read the token frames, so its success is unchanged by construction. (b) Successful episodes of two tasks, three frames each (start, grasp, placement): top, ``pick up the black bowl from table center and place it on the plate''; bottom, ``pick up the black bowl next to the ramekin and place it on the plate''.}
\label{fig:libero}
\end{figure*}

The equivariance constraint makes the command independent of the reference frame chosen for each token (Section~\ref{sec:invariance}). We test two consequences commonly credited to equivariant policies: robustness to a change of frame, and a lower number of demonstrations, since the constraint supplies the behaviour in other frames that a network would otherwise learn from data~\cite{yang2024equibot,ryu2023diffusionedf}.

\begin{table}[t]
\centering
\caption{Reorientation: ablations of the equivariance constraint and of the attention mask.
Success (\%), mean $\pm$ SD. Frame change: every link's reference frame
changed, policy unchanged, worst of three conventions. With two layers, the adjacency mask
reaches only bodies within two edges.}
\label{tab:reorient}
\footnotesize
\setlength{\tabcolsep}{3pt}
\begin{tabular}{lccc}
\toprule
Network & Mechanism & Canonical & Frame change \\
\midrule
Unconstrained & \cite{xmop2024,rodrinet2025} & 97.7 $\pm$ 2.5 & 46.0 $\pm$ 45.5 \\
Full attention & \cite{gupta2022metamorph} & 98.1 $\pm$ 1.7 & 97.7 $\pm$ 1.8 \\
Adjacency mask & \cite{sferrazza2024body} & 75.5 $\pm$ 6.2 & 69.0 $\pm$ 6.8 \\
Random mask & --- & 51.9 $\pm$ 28.9 & 49.7 $\pm$ 28.4 \\
\midrule
\textbf{Screw Attention} & --- & \textbf{99.2 $\pm$ 0.3} & \textbf{98.6 $\pm$ 0.6} \\
\bottomrule
\end{tabular}
\end{table}

For the first, we change the frame of every link at evaluation time, without retraining. On reorientation (Table~\ref{tab:reorient}), Screw Attention and the three variants that keep the constraint lose under 7 points. Their small losses come from the three pair features that describe where the frames are placed, which this test recomputes (Appendix~\ref{app:features}). The unconstrained network, whose weights mix the six spatial components, loses about half of its success, although it matches Screw Attention canonically. It has fitted the frame convention of its training data. On LIBERO-Spatial (Fig.~\ref{fig:libero}a), the frame of every token is rotated about its origin while the pair features keep their canonical values, so the test isolates the transported messages. Screw Attention is unchanged within one binomial standard error, as the construction predicts, while every learned baseline falls below 3\%.

\begin{figure}[t]
\centering
\begin{tikzpicture}
\begin{axis}[
  width=\columnwidth, height=4.6cm,
  xmode=log, log basis x=10,
  xmin=20, xmax=1000, ymin=40, ymax=100,
  xtick={25,50,100,800}, xticklabels={25,50,100,800},
  ytick={40,50,60,70,80,90,100},
  xlabel={Demonstrations}, ylabel={Success (\%)},
  grid=major, grid style={gray!20},
  tick label style={font=\footnotesize}, label style={font=\footnotesize},
  legend style={font=\scriptsize, at={(0.97,0.05)}, anchor=south east, draw=none,
                fill=white, fill opacity=0.85, text opacity=1},
]
\addplot[lane-g, thick, mark=*, mark size=1.6pt]
  coordinates {(25,57.2) (50,66.8) (100,82.7) (800,98.4)};
\addlegendentry{Screw Attention}
\addplot[lane-p, thick, dashed, mark=square*, mark size=1.6pt]
  coordinates {(25,57.6) (50,69.8) (100,81.2) (800,98.3)};
\addlegendentry{Unconstrained}
\addplot[lane-s, thick, dotted, mark=triangle*, mark size=1.9pt]
  coordinates {(25,43.9) (50,51.9) (100,75.8) (800,93.1)};
\addlegendentry{Full attention}
\end{axis}
\end{tikzpicture}
\caption{Reorientation success against the number of demonstrations (behaviour cloning only, without the PPO stage of Tables~\ref{tab:results} and~\ref{tab:reorient}).}
\label{fig:budget}
\end{figure}
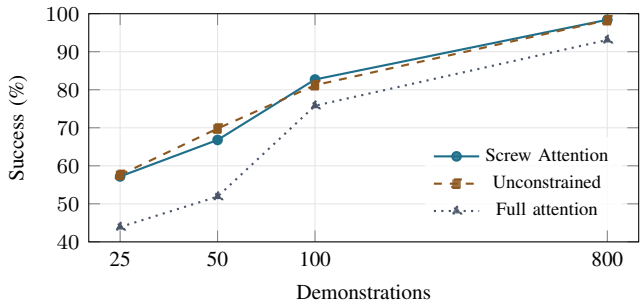

The second consequence does not hold here. With behaviour cloning on 25 to 800 demonstrations (Fig.~\ref{fig:budget}), Screw Attention and the unconstrained network learn at the same rate. All demonstrations share one frame convention, so there is no variation in the data for the constraint to spare. The attention mask does reduce the number of demonstrations needed: full attention trails at small budgets, which suggests that the mask spares the network from learning which tokens to ignore. After PPO, the choice of mask also matters. The adjacency mask and a random mask of the same density fall well below both full attention and the ancestor and descendant masks of Screw Attention (Table~\ref{tab:reorient}).

\subsection{LIBERO-Spatial}
\label{sec:libero}

LIBERO-Spatial tests the representation on a standard benchmark, against baselines that reimplement published mechanisms at the same parameter budget (Fig.~\ref{fig:libero}a). One policy is fitted to all ten tasks, and Fig.~\ref{fig:libero}b shows successful episodes in which it emits every action with no analytic controller underneath. Screw Attention has the highest success. The baselines separate three ingredients. Grouping the observation into tokens brings the largest gain (transformer against flat MLP). Giving the relation as edge features adds a smaller gain, within the spread of the two networks (graph network against transformer). Transporting the messages with the relation adds 12.6 points more (Screw Attention against the graph network; $p < 10^{-10}$, Fisher's exact test). The flat MLP needs 27 times more parameters to approach Screw Attention, and still stays below it ($p = 0.018$). At a matched budget, the geometric representation is therefore far more efficient than learning the relations from flat inputs. It is the slowest network at inference, but one forward pass still takes less than 1\,ms, far below the 50\,ms control period.

When every object is displaced by up to 5\,cm, all networks lose 6 to 10 points, except the matched flat MLP, whose canonical success is already low. The training data also vary the object positions, so this test measures a harder distribution, not robustness. Appendix~\ref{app:robust} compares a perturbed initial robot configuration with published image-based models, and Appendix~\ref{app:fit} compares training fit with success.

\begin{figure}[t]
\centering
\begin{tikzpicture}
\begin{groupplot}[
  group style={group size=2 by 1, horizontal sep=0.25cm, y descriptions at=edge left},
  scale only axis, width=3.55cm, height=3.6cm,
  xmin=-1.5, xmax=43, ymin=-90, ymax=8,
  xtick={0,10,20,40}, ytick={-80,-60,-40,-20,0},
  ylabel={Drop from canonical (points)},
  grid=major, grid style={gray!20},
  title style={font=\footnotesize, yshift=-4pt},
  tick label style={font=\footnotesize}, label style={font=\footnotesize},
]
\nextgroupplot[title={(a) Pose noise}, xlabel={Pose noise $\sigma$ (mm)},
  legend columns=2, legend cell align=left, legend to name=robustlegend,
  legend style={font=\scriptsize, draw=gray!50, fill=white, inner sep=2pt,
                /tikz/every even column/.append style={column sep=4pt}}]
\addplot[black, thick, dashed, mark=star, mark size=2.4pt] coordinates {(0,0) (20,-50.8) (40,-80.8)};
\addlegendentry{Analytic controller}
\addplot[lane-p, mark=square*, mark size=1.5pt] coordinates {(0,0) (20,-1.5) (40,-40.2)};
\addlegendentry{Graph network}
\addplot[lane-s, dashed, mark=triangle*, mark size=1.9pt] coordinates {(0,0) (20,-3.8) (40,-32.7)};
\addlegendentry{Transformer}
\addplot[gray, dotted, thick, mark=diamond*, mark size=1.9pt] coordinates {(0,0) (20,-6.1) (40,-32.9)};
\addlegendentry{Flat MLP, 438k}
\addplot[lane-g, very thick, mark=*, mark size=1.6pt]
  coordinates {(0,0) (5,0.0) (10,-0.8) (20,-3.4) (40,-34.3)};
\addlegendentry{Screw Attention}

\nextgroupplot[title={(b) Joint offset}, xlabel={Hand error (mm)}]
\addplot[black, thick, dashed, mark=star, mark size=2.4pt]
  coordinates {(0,0) (5.1,-0.6) (9.9,-6.0) (19.8,-20.8) (40.6,-49.0)};
\addplot[lane-g, very thick, mark=*, mark size=1.6pt]
  coordinates {(0,0) (5.1,-0.4) (9.9,-1.7) (19.8,-13.9) (40.6,-42.5)};
\end{groupplot}
\node[anchor=center] at ($(group c1r1.south east)!0.5!(group c2r1.south west) + (0,0.73cm)$)
  {\pgfplotslegendfromname{robustlegend}};
\end{tikzpicture}
\caption{LIBERO-Spatial under (a) pose noise and (b) a constant joint-angle offset, policies
unchanged: drop from canonical success. In (b), offsets of 0.25, 0.5, 1 and 2$^\circ$ (standard
deviation per joint) displace the hand by 5, 10, 20 and 41\,mm on average. Lines: mean per network.}
\label{fig:robust}
\end{figure}
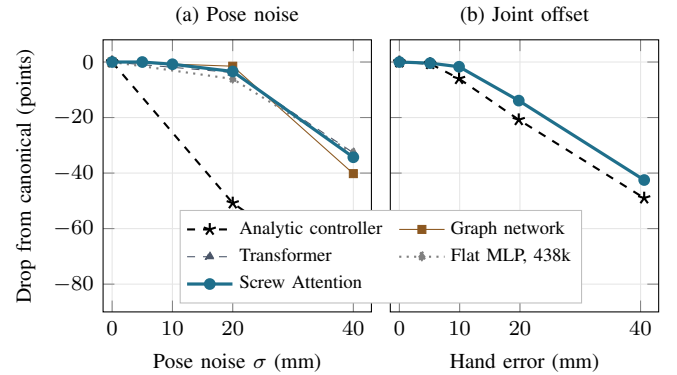

Since the object poses are read from the simulator (Section~\ref{sec:tasks}), we emulated the error of a pose estimator. We added Gaussian noise of standard deviation $\sigma$ to the position of every object in the observation, independently at every step, while the simulator kept the true poses (Fig.~\ref{fig:robust}a). All learned networks lose only a few points up to 20\,mm, about seven times the largest median translation error reported for FoundationPose~\cite{wen2024foundationpose} on industrial components~\cite{masuhr2025foundationpose}. The analytic controller, given the same poses, loses more than half of its success. Pose noise therefore separates the learned policies from the analytic pipeline, but not from each other.

\subsection{Joint offsets}
\label{sec:joint}
The relative poses and joint screws of the robot tokens are computed from the robot's kinematic model and its joint angles, and so is the hand pose used by the target relations. On a real robot, this model is often only approximately calibrated. To emulate calibration error, we added a constant offset to every arm joint angle in the observation, drawn once per episode from a Gaussian distribution, while the simulator kept the true angles. The analytic controller received the same corrupted state. The factory kinematic model of Franka arms places the hand 3.3 to 8.8\,mm from its true position on average~\cite{franzese2025mukca}, which corresponds to offsets of about 0.2 to 0.4$^\circ$.

At offsets of 0.25 and 0.5$^\circ$, comparable to this range (Fig.~\ref{fig:robust}b), Screw Attention loses less than 2 points, against up to 6 for the analytic controller. The learned policy therefore tolerates a factory-calibrated Franka arm. Larger offsets degrade both sharply. Under both pose noise and joint offsets, Screw Attention degrades less than the analytic controller given the same inputs. It therefore requires no more precise calibration than the analytic pipeline commonly used on such arms.

\section{Discussion}
\label{sec:discussion}
\label{sec:doesnot}

Analysing the results across tasks identifies the configurations in which the geometric structure of Screw Attention is beneficial (and where it is not), and clarifies which property of the network produces the gains.

\subsection{When the geometry contributes}

The geometry contributes far more on the affordance task than on the insertion task, although both tasks use the same number of tokens. The difference lies in what the rest of the system already provides. In the insertion task, the analytic controller is part of the network
through the model term (Eq.~\ref{eq:residual}), so the geometry required by the task is supplied correctly at every step. This likely explains why corrupting the policy's own copy of that geometry has little effect. In the affordance task, by contrast, nothing in the loop encodes which candidate is reachable. The policy must infer reachability from the relation between the arm and each candidate, and corrupting that relation costs 73.9 points. The geometric structure helps when it carries information essential to the task that no other part of the system provides. 
Appendix~\ref{app:nulls} shows two task designs that
fall into the same category as the insertion task: they need only one position vector per object or per link,
which the scalar features already supply.

\subsection{What the equivariance constraint provides}

The equivariance constraint provides robustness to how the robot is described. Under a frame change, the constrained networks retain nearly all of their success, whereas the unconstrained network loses about half on average and the LIBERO baselines lose nearly all. Although the equivariance constraint does not reduce the number of demonstrations needed, it changes what the network learns. An unconstrained network can reach the same success by fitting the frame convention of its training data, a shortcut that fails when the convention changes. The attention mask is what reduces the number of demonstrations needed. 

Among the geometric perturbations on LIBERO-Spatial, the frame change is the one that separates the networks. Pose noise and object displacement reduce the success of all learned networks by similar amounts, and Screw Attention keeps the lead it has on the canonical episodes. In Screw Attention, this robustness follows from the construction and requires none of the augmentation over frame conventions that other methods need (Section~\ref{sec:related}). A network that is equivariant only to a global frame change, such as GATr~\cite{brehmer2023gatr}, does not have this robustness by construction. Redefining the frame of each link independently changes the relative transforms between links. No single global frame change can reproduce these independent per-link redefinitions, so such local changes lie outside the symmetry handled by such architectures (Appendix~\ref{app:gatr}).

\subsection{Why the learned policies tolerate observation errors}
Under pose noise and joint-angle offsets, Screw Attention loses less success than the analytic controller that receives the same corrupted observations (Fig.~\ref{fig:robust}). The analytic controller relies on precise, discrete decisions, such as closing the gripper only when the hand is within a few millimetres of its target. A small observation error can then trigger such a decision at the wrong moment or prevent it altogether. The learned policies, by contrast, map the observation continuously to commands. A small error shifts the command without altering the overall course of the task. Under pose noise, this tolerance is shared by all learned networks. It disappears once the error exceeds what a grasp can absorb.

\section{Limitations}
\label{sec:limitations}

As this study focused on the representation rather than on its deployment, Screw Attention was validated only in simulation. In this setting, the policies read object poses from the simulator. A deployment therefore requires a pose estimator. We emulated the pose estimation errors by adding translation noise to the object poses (Section~\ref{sec:libero}), but rotation errors were not tested. Similarly, photometric factors, such as camera placement, lighting, background and texture, are outside the scope of this paper, since they reach the policy only through the pose estimator. Validating the approach on a real robot, with an estimator in the loop, is one of the main directions for future work.

Additionally, because the rigid-body algebra is inside the network, Screw Attention depends on the accuracy of the robot model. The pair terms and inertias come from it, and the model errors were tested only as joint offsets (Section~\ref{sec:joint}), not as errors in link lengths or inertias. In our tests, this dependence is weaker than for the analytic controller.

Finally, our construction assumes that each body is driven by one revolute or prismatic joint, which holds for the fixed-base serial arms studied here but excludes bodies with several joints  and floating bases (mobile manipulators, legged robots). Screw Attention is also the slowest of the compared networks at inference, since it transforms the geometric channels of every ordered pair of tokens, a cost that grows quadratically with the number of tokens. It nevertheless remains far below the control period on the tasks studied here.

\section{Conclusion}
\label{sec:conclusion}

We introduced Screw Attention, an architecture that integrates rigid-body algebra inside a transformer layer. With robot bodies and scene objects as input tokens, the relative pose and the joint screw of each pair of bodies transform the messages exchanged between tokens, while the attention weights see only frame-invariant quantities. Instead of learning the geometry from data, the network is constrained to respect it by construction. This is achieved by integrating geometric quantities provided by the robot model. The learned weights mix only their channels, never their spatial components.
The robot's analytic model thus becomes an input to the network. When an analytic controller is available, the network can also learn a residual on its command, weighted by a learned gate.
With this construction, a single transformer layer can express the velocity recursion along the kinematic chain, and its command does not depend on how frames are attached to links.  On four tasks, Screw Attention matched or outperformed other learned networks of the same size and exceeded the analytic controller on three of them. On LIBERO-Spatial, a policy with 16{,}162 parameters reaches 97.3\%, above networks of the same size and above a flat network with 27 times more parameters. We showed that the geometric structure is most beneficial when the task depends on relations between frames that no other part of the system supplies. On an affordance task, correct relations are worth 73.9 points over shuffled ones. On an insertion task, where the analytic controller already supplies the geometry, the margin is small, but the learned residual adds 17.3 points to that controller. The equivariance constraint makes the policy robust to how the robot is described, and the attention mask reduces the number of demonstrations needed. On the reorientation task, a frame change costs Screw Attention 0.6 points  while a policy without this constraint loses 51.7 points. Moreover, Screw Attention tolerates pose noise up to 10\,mm and joint offsets within the factory calibration of a Franka arm. Future work will validate the approach on a real robot with a pose estimator in the loop, and extend the construction to bodies driven by several joints and to moving bases, which would open it to legged robots, musculoskeletal models and other articulated rigid-body systems.

\bibliographystyle{IEEEtran}
\bibliography{refs}

\clearpage
\makeatletter
\setlength{\@fptop}{0pt}
\setlength{\@fpsep}{12pt}
\setlength{\@fpbot}{0pt plus 1fil}
\setlength{\@dblfptop}{0pt}
\setlength{\@dblfpsep}{12pt}
\setlength{\@dblfpbot}{0pt plus 1fil}
\makeatother
\appendices

\twocolumn[{
\centering
\captionof{table}{Training settings. BC: behaviour cloning. Each PPO update collects
256 environments $\times$ 32 steps. Readout normalisation divides each pairing term of the readout by its root-mean-square over joints and channels. Rewards: $\theta$, angle between the end-effector orientation and the goal; $\mathbf{a}$, action; $e_k = \max(\delta_p/0.04\,\mathrm{m},\ \delta_r/0.35\,\mathrm{rad})$, error to candidate $k$ from the position and orientation errors $\delta_p$ and $\delta_r$, reached when $e_k < 1$; $c$, penalty quadratic in how far the clearance of each link to the object falls below 8\,cm (weight 2); the touch penalty rises from 20\% to its full value over the first 400 updates; $e$, lateral offset of the peg tip from the socket axis; $d$, its depth; $d^\ast = 30$\,mm, inserted when $d > d^\ast$ and $e$ is within the half-width of the hole.}
\label{tab:training}
\footnotesize
\setlength{\tabcolsep}{3pt}
\begin{tabular}{lcccc}
\toprule
& Reorientation & Affordance & Insertion & LIBERO \\
\midrule
BC demonstrations & 400 episodes & 1{,}500 episodes & --- & 500 episodes \\
BC steps & 3{,}000 & 20{,}000 & --- & 144{,}000 \\
BC rate & $3\times10^{-3}$ & $3\times10^{-3}$ & --- & $10^{-3} \to 10^{-4}$ \\
PPO updates & 400 & 1{,}200 & 1{,}200 & --- \\
PPO rate & $3\times10^{-4}$ & $3\times10^{-4} \to 3\times10^{-5}$ & $3\times10^{-4} \to 3\times10^{-5}$ & --- \\
Entropy coefficient & $10^{-3}$ & $2\times10^{-3}$ & $10^{-3}$ & --- \\
Target KL & 0.02 & 0.02 & 0.02 & --- \\
Readout normalisation & No & Yes & No & No \\
Reward per step & $-\theta - 10^{-4}\lVert\mathbf{a}\rVert^2 - 10^{-3}\lVert\dot{\mathbf{q}}\rVert^2$ & $-\mathrm{softmin}_k e_k - c - 10^{-4}\lVert\mathbf{a}\rVert^2$ & $-20e - 10\max(d^\ast - d, 0) - 10^{-4}\lVert\mathbf{a}\rVert^2$ & --- \\
Reward on outcome & --- & $+5$ reached, $-25$ touch, once & $+1$ per step once inserted & --- \\
Evaluation & 512 episodes & 1{,}024 scenes & 768 episodes & 500 episodes \\
\bottomrule
\end{tabular}
\vspace{1.5em}
}]

\section{Training details}
\label{app:training}

Table~\ref{tab:training} lists the training settings. All networks were implemented in JAX, and the tasks were simulated in MuJoCo~\cite{todorov2012mujoco}. On the robot tasks, PPO used a clipping ratio of 0.2, a discount of 0.99, a GAE parameter of 0.95, four epochs of four minibatches per update, and a critic with its own optimiser. Each run used a single NVIDIA L4 GPU. Behaviour cloning was not used on the insertion task, since the model term already implements the demonstrator and the residual target is zero. The LIBERO networks were fitted with Adam, with the gradient norm clipped at 1, to the benchmark's demonstrations re-executed with each object offset by up to 5\,cm (500 demonstrations).

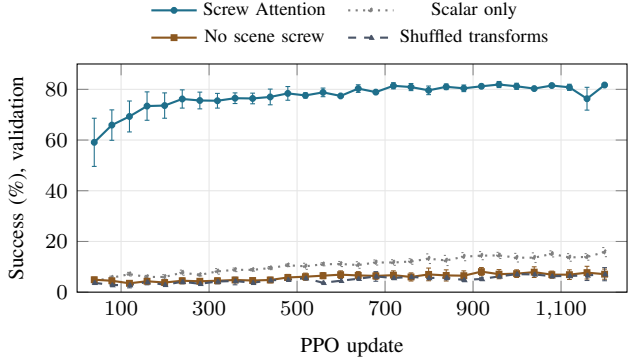
\begin{figure}[tb]
\centering
\begin{tikzpicture}
\begin{axis}[
  width=\columnwidth, height=4.6cm,
  xlabel={PPO update}, ylabel={Success (\%), validation},
  xmin=0, xmax=1250, ymin=0, ymax=90,
  xtick={100,300,500,700,900,1100}, ytick={0,20,40,60,80},
  grid=major, grid style={gray!20},
  tick label style={font=\footnotesize}, label style={font=\footnotesize},
  legend columns=2,
  legend style={font=\scriptsize, at={(0.5,1.03)}, anchor=south, draw=none,
                /tikz/every even column/.append style={column sep=6pt}},
  every axis plot/.append style={thick, mark size=0.9pt,
    error bars/error bar style={thin}},
]
\addplot[lane-g, mark=*, error bars/.cd, y dir=both, y explicit] coordinates {
  (39,59.1) +- (0,9.5)   (79,65.9) +- (0,6.0)   (119,69.3) +- (0,6.1)  (159,73.4) +- (0,5.6)
  (199,73.6) +- (0,5.0)  (239,76.2) +- (0,3.6)  (279,75.6) +- (0,3.3)  (319,75.5) +- (0,2.9)
  (359,76.5) +- (0,2.1)  (399,76.4) +- (0,2.1)  (439,77.0) +- (0,3.1)  (479,78.4) +- (0,2.7)
  (519,77.6) +- (0,1.0)  (559,78.8) +- (0,1.7)  (599,77.4) +- (0,0.6)  (639,80.3) +- (0,1.5)
  (679,78.9) +- (0,0.7)  (719,81.4) +- (0,1.2)  (759,80.9) +- (0,1.4)  (799,79.6) +- (0,1.7)
  (839,81.0) +- (0,1.1)  (879,80.4) +- (0,1.3)  (919,81.2) +- (0,0.8)  (959,81.9) +- (0,1.2)
  (999,81.2) +- (0,1.2)  (1039,80.3) +- (0,0.5) (1079,81.5) +- (0,0.7) (1119,80.8) +- (0,1.2)
  (1159,76.3) +- (0,4.5) (1199,81.7) +- (0,0.7)};
\addlegendentry{Screw Attention}
\addplot[gray, dotted, mark=diamond*, error bars/.cd, y dir=both, y explicit] coordinates {
  (39,4.4) +- (0,0.6)  (79,5.7) +- (0,0.6)  (119,7.1) +- (0,0.7)  (159,6.0) +- (0,0.6)
  (199,6.0) +- (0,0.9)  (239,7.7) +- (0,1.0)  (279,6.8) +- (0,0.4)  (319,8.2) +- (0,1.1)
  (359,8.8) +- (0,0.7)  (399,8.8) +- (0,0.5)  (439,9.5) +- (0,0.6)  (479,10.6) +- (0,0.5)
  (519,10.2) +- (0,1.2)  (559,11.0) +- (0,0.6)  (599,11.1) +- (0,1.0)  (639,10.7) +- (0,1.3)
  (679,11.7) +- (0,1.0)  (719,11.7) +- (0,1.1)  (759,12.1) +- (0,1.1)  (799,13.2) +- (0,1.9)
  (839,12.5) +- (0,2.0)  (879,14.0) +- (0,1.6)  (919,14.4) +- (0,1.6)  (959,14.5) +- (0,1.2)
  (999,13.6) +- (0,1.7)  (1039,13.6) +- (0,1.9)  (1079,15.1) +- (0,1.2)  (1119,13.7) +- (0,1.5)
  (1159,13.9) +- (0,1.3)  (1199,15.8) +- (0,1.8)};
\addlegendentry{Scalar only}
\addplot[lane-p, mark=square*, error bars/.cd, y dir=both, y explicit] coordinates {
  (39,4.9) +- (0,1.0)  (79,4.5) +- (0,0.7)  (119,3.5) +- (0,0.6)  (159,4.3) +- (0,0.9)
  (199,3.8) +- (0,0.9)  (239,4.5) +- (0,1.1)  (279,4.3) +- (0,0.6)  (319,4.5) +- (0,1.1)
  (359,4.8) +- (0,0.9)  (399,4.6) +- (0,0.9)  (439,4.8) +- (0,1.1)  (479,5.8) +- (0,1.2)
  (519,6.1) +- (0,1.4)  (559,6.5) +- (0,1.2)  (599,6.9) +- (0,1.5)  (639,6.6) +- (0,1.5)
  (679,6.4) +- (0,1.3)  (719,6.7) +- (0,1.7)  (759,6.0) +- (0,1.6)  (799,7.0) +- (0,2.5)
  (839,6.6) +- (0,2.2)  (879,6.5) +- (0,1.9)  (919,8.1) +- (0,1.6)  (959,7.1) +- (0,1.8)
  (999,7.3) +- (0,1.5)  (1039,7.9) +- (0,2.1)  (1079,6.9) +- (0,1.4)  (1119,6.9) +- (0,1.9)
  (1159,7.7) +- (0,2.5)  (1199,7.1) +- (0,2.0)};
\addlegendentry{No scene screw}
\addplot[lane-s, dashed, mark=triangle*, error bars/.cd, y dir=both, y explicit] coordinates {
  (39,3.7) +- (0,0.6)  (79,2.8) +- (0,0.3)  (119,2.8) +- (0,1.1)  (159,4.0) +- (0,1.1)
  (199,2.9) +- (0,0.0)  (239,4.1) +- (0,0.8)  (279,3.3) +- (0,0.2)  (319,4.2) +- (0,0.9)
  (359,4.4) +- (0,1.5)  (399,3.7) +- (0,0.0)  (439,4.7) +- (0,0.6)  (479,4.9) +- (0,0.0)
  (519,5.5) +- (0,0.4)  (559,3.7) +- (0,0.0)  (599,4.5) +- (0,0.0)  (639,5.3) +- (0,0.0)
  (679,6.2) +- (0,1.9)  (719,5.6) +- (0,0.3)  (759,6.0) +- (0,0.3)  (799,5.6) +- (0,0.3)
  (839,5.7) +- (0,0.8)  (879,4.8) +- (0,0.5)  (919,5.2) +- (0,0.5)  (959,6.2) +- (0,0.1)
  (999,7.1) +- (0,0.7)  (1039,7.0) +- (0,0.2)  (1079,6.3) +- (0,0.5)  (1119,6.4) +- (0,0.6)
  (1159,6.6) +- (0,2.0)  (1199,7.1) +- (0,2.6)};
\addlegendentry{Shuffled transforms}
\end{axis}
\end{tikzpicture}
\caption{Affordance task: validation success (512 fixed scenes) during PPO, mean
$\pm$ standard error.}
\label{fig:curves}
\end{figure}

\textbf{Learning curves.} Fig.~\ref{fig:curves} shows the validation success on the
affordance task during PPO. Screw Attention is above every ablation from the first evaluation
and levels off near 80\% after about 700 updates. The ablations are still improving at the end
of training: scalar only rises from 4\% to 16\%, and the networks without the scene screw or
with shuffled transforms rise from 4\% to 7\%.

\section{Verification of the construction}
\label{sec:verification}

This appendix describes the numerical checks of the constructions in
Sections~\ref{sec:representability} and~\ref{sec:invariance}.

\textbf{Spatial algebra.} The relative transforms, motion subspaces and spatial inertias were
compared with those of an independent rigid-body engine, MuJoCo~\cite{todorov2012mujoco}, on
two robot arms and one synthetic model. All agreed to $10^{-13}$ or better.

\textbf{Velocity recursion.} With the weights set as in Section~\ref{sec:representability}, a single layer reproduced the velocity recursion of Eq.~\ref{eq:velocity} to within $4\times10^{-15}$ on five kinematic trees. The construction sets three weights and leaves every other weight at zero, whatever the width of the layer.

\textbf{Jacobian-transpose law.} With the weights of the inward sweep set by hand, a wrench applied to the end-effector token was read out as the joint torques $\boldsymbol{\tau} = \mathbf{J}^{\top}\mathbf{F}$ to within $10^{-9}$, on a chain, a branched tree and a tree with welded bodies.

\textbf{Frame change.} The per-body frames were changed by random rigid transforms. Three of the pair features depend on where the frames are placed on the bodies (Appendix~\ref{app:features}), so the test was run in two parts. With these features removed, the whole layer, attention included, commuted with the frame change. With the attention
weights held fixed, the transported update commuted with it for any pair features. Both held to within $10^{-10}$. A layer in which $\mathbf{W}_g$ mixes the six spatial components failed both tests. A further test verifies that no learned parameter acts on the six spatial components.

\textbf{Common frame.} Choosing the base frame as the frame of every token,
$\mathbf{T}_i = \X_{0i}$ in Eq.~\ref{eq:frame-change}, turns every relative transform into the identity and expresses every quantity in one common frame. With the frame-dependent pair features removed, the layer gave the same torques in this common frame as in the original
frames, to within $10^{-9}$ on five kinematic trees, with both the motion readout and the force readout with the inertia coupling. Appendix~\ref{app:gatr} discusses what this correspondence implies for geometric-algebra transformers.

\begin{table}[t]
\centering
\caption{The fourteen entries of $\boldsymbol{\varphi}_{ij}$, in three groups.}
\label{tab:features}
\footnotesize
\setlength{\tabcolsep}{3pt}
\begin{tabular}{cll}
\toprule
Entries & Feature & Value \\
\midrule
\multicolumn{3}{l}{\emph{Relative placement of the two frames}} \\
1 & Distance & $\lVert \mathbf{r}_{ij} \rVert$ \\
2 & Inverse distance & $1/(1 + \lVert \mathbf{r}_{ij} \rVert)$ \\
3 & Rotation cosine & $(\operatorname{tr}\mathbf{E}_{ij} - 1)/2$ \\
\midrule
\multicolumn{3}{l}{\emph{Relation of the two joint axes (Eq.~\ref{eq:pair-axes})}} \\
4 & First pairing & $\kappa(\mathbf{S}_i, \X_{ij}\mathbf{S}_j)$ \\
5 & Second pairing & $\lambda(\mathbf{S}_i, \X_{ij}\mathbf{S}_j)$ \\
\midrule
\multicolumn{3}{l}{\emph{Structure of the kinematic tree}} \\
6--7 & Joint type of $j$ & Revolute, prismatic (indicators) \\
8 & Graph distance & $\min(h_{ij}, 8)/8$ \\
9--13 & Relation of $j$ to $i$ & Self, ancestor, descendant, sibling, \\
 & & unrelated (indicators) \\
14 & Joint of $j$ moves $i$ & Indicator \\
\bottomrule
\end{tabular}
\end{table}

\section{Relation to geometric-algebra transformers}
\label{app:gatr}

GATr~\cite{brehmer2023gatr} builds $E(3)$-equivariant attention by representing every quantity as a multivector of the projective geometric algebra and combining multivectors with the geometric product. Screw Attention relates to it through its algebra and through its symmetry.

\textbf{Algebra.} Our geometric channels are the bivectors of that algebra, which form the Lie algebra $\mathfrak{se}(3)$ of screws under the commutator. A twist is a bivector, a grade-2 element of the algebra. A spatial transform $\X_{ij}$ is the adjoint action of a motor, the element of the algebra that represents a rigid displacement, on screws. Finally, the pairings $\kappa$ and $\lambda$ of Eq.~\ref{eq:pairings} are the frame-invariant bilinear forms, which map two screws to a number and are unchanged when both are moved to another frame. We keep only the grade that rigid-body mechanics uses, and Section~\ref{sec:representability} shows that this restriction suffices for the two computations considered there: a single layer reproduces the
velocity recursion and the Jacobian-transpose law.

\textbf{Symmetry.} GATr is equivariant to a change of the one common frame of the whole scene, whereas each of our tokens carries its own frame, changed independently by the transforms $\mathbf{T}_i$ of Eq.~\ref{eq:frame-change}. In the common-frame representation of Appendix~\ref{sec:verification}, GATr's symmetry is the
diagonal subgroup of our per-token gauge, in which every $\mathbf{T}_i$ is the same transform (without reflections, which $E(3)$ includes). A model equivariant only to this subgroup is not invariant to re-describing each link independently, since that changes the relative transforms $\X_{ij}$ in a way no change of the common frame can reproduce. Such a model must learn the invariance from data. Ours is invariant by construction, up to the three frame-dependent pair features of Appendix~\ref{app:features}. This holds only for the torque readouts. The twist readout is invariant to rotations only (Section~\ref{sec:invariance}).

Screw Attention therefore uses a smaller algebra than GATr and a larger symmetry group, with one frame per token.

\section{Terms of the attention score}
\label{app:features}

\begin{table}[t]
\centering
\caption{Baselines and ablations, and the corresponding published mechanisms.}
\label{tab:correspondence}
\footnotesize
\setlength{\tabcolsep}{3pt}
\begin{tabular}{l>{\raggedright\arraybackslash}p{3.0cm}cc}
\toprule
Method & Mechanism it instantiates & Used on & Params \\
\midrule
Transformer & Attention over body tokens~\cite{sferrazza2024body} & LIBERO & 17{,}350 \\
Graph network & Message passing on the kinematic graph~\cite{abdnet2026} & LIBERO & 17{,}038 \\
Flat MLP, 17k & No structure & LIBERO & 17{,}572 \\
Flat MLP, 438k & No structure & LIBERO & 438{,}384 \\
Unconstrained & Learned matrix on the geometric components~\cite{xmop2024,rodrinet2025} & Reorientation & --- \\
Scalar only & Joint attributes as score features~\cite{morphformer2026} & Robot tasks & --- \\
Shuffled transforms & Correct transforms on the wrong pairs; on affordance, robot--scene pairs only & Robot tasks & --- \\
Full attention & Attention without a mask~\cite{gupta2022metamorph} & Reorientation & --- \\
Adjacency mask & Adjacency in the kinematic tree~\cite{sferrazza2024body} & Reorientation & --- \\
Random mask & Density-matched random mask & Reorientation & --- \\
No scene screw & Approach axis removed from the candidate tokens & Affordance & --- \\
Shuffled inertias & Correct spatial inertias on the wrong bodies & Insertion & --- \\
Flat MLP, 17k & No structure & Affordance & 17{,}752 \\
Flat MLP, 405k & No structure & Affordance & 405{,}000 \\
Analytic controller & The classical pipeline, given the same poses & All & 0 \\
\midrule
\textbf{Screw Attention} & Relative pose and joint screw on the value path & All & \hd{16{,}162--\\19{,}122} \\
\bottomrule
\end{tabular}
\end{table}

\begin{table}[tb]
\centering
\caption{LIBERO-Spatial, perturbed initial robot configuration: success (\%) and drop from canonical success (points). Only the drops are comparable across rows. Screw Attention: a single policy, whose canonical success differs from the mean of Fig.~\ref{fig:libero}a.}
\label{tab:robotinit}
\footnotesize
\setlength{\tabcolsep}{3pt}
\begin{tabular}{lcccc}
\toprule
Policy & Input & Canonical & Perturbed & Drop \\
\midrule
MINERVA~\cite{minerva2026} & Images & 95.75 & 57.1 & $-38.7$ \\
UniVLA~\cite{univla2025} & Images & 95.2 & 50.3 & $-44.9$ \\
\midrule
Analytic controller & Poses & 81.6 & 81.3 & $-0.3$ \\
Screw Attention & Poses & 94.0 & 87.2 & $-6.8$ \\
\bottomrule
\end{tabular}
\end{table}

Table~\ref{tab:features} lists the fourteen entries of the pair features
$\boldsymbol{\varphi}_{ij}$ in Eq.~\ref{eq:scores}. We denote by $\mathbf{r}_{ij}$ the position of the origin of frame $j$ in frame $i$, by $\mathbf{E}_{ij}$ the rotation from frame $j$ to frame $i$, and by $h_{ij}$ the number of edges between bodies $i$ and $j$ in the kinematic tree. For a pair involving an object, the joint entries are zero, the graph distance
takes its maximum value, and the relation is \emph{unrelated}, except for the pair formed by the object with itself. Entries 1 to 3 depend on where the frames are placed on the bodies and therefore change under the frame change of Section~\ref{sec:invariance}. The other eleven
entries do not. The attention weights depend on the frames only through entries 1 to 3.

\section{Baselines, ablations and the mechanisms they instantiate}
\label{app:controls}

We could not compare Screw Attention with results reported for published policies, because each was obtained with a different task, parameter budget, dataset and evaluation protocol. Instead, each baseline or ablation reimplements the mechanism of a published line of work at a matched parameter budget, and is trained on the same data and with the same configuration as Screw Attention.
Table~\ref{tab:correspondence} lists the baselines and ablations, the mechanism each instantiates and the experiments in which it is used. Each therefore represents a design choice, not a full published method. In one case, the baseline is better informed than the method it represents: the graph network, in LIBERO-Spatial, receives the true relative transforms as edge features, whereas ABD-NET~\cite{abdnet2026} learns its edge features and inertias. In the same task, the three baselines fitted the training data better than Screw Attention (Appendix~\ref{app:fit}), so their lower success does not come from a worse fit.

\textbf{Scope.} Screw Attention consumes object poses, not images. Our results are therefore not directly comparable with those of image-based policies on the same benchmark. We therefore compare only drops from canonical success, never success rates. In all experiments the poses were read from the simulator, with noise added where stated. A deployment would require a pose estimator, whose parameters are not included in the counts reported here.

\textbf{Shuffled transforms on the affordance task.} The affordance policy uses the force readout, in which each joint torque is computed from forces transported along the robot's own transforms. Permuting all transforms, as on the other tasks, corrupts the mapping from forces to torques in addition to the relations. With that ablation, the network could not
imitate the demonstrator. On this task specifically, the ablation permuted only the transforms from the robot tokens to the scene tokens, at every step. The robot's own transforms remained correct, the network fitted the demonstrations as well as the other networks, and only the relations to the object and the candidates were wrong.

\section{Robustness on LIBERO-Spatial}
\label{app:robust}

Table~\ref{tab:robotinit} reports the drop under a perturbed initial
robot configuration, generated as in LIBERO-plus~\cite{liberoplus2025}, together with the
drops reported for two image-based baselines. We compare drops instead of success rates, because
those models use images and are evaluated on four suites, whereas our policy uses poses and is
evaluated on LIBERO-Spatial. Screw Attention loses 6.8 points, against 38.7 for MINERVA and
44.9 for UniVLA. The perturbation moves only the robot, and the objects remain where the
training data placed them, so it is expected to affect a pose-based policy less.

\section{Training fit on LIBERO-Spatial}
\label{app:fit}

Table~\ref{tab:fit} compares each network's fit to the training data with its canonical
success on LIBERO-Spatial. The three baselines fitted the training data better than
Screw Attention: the large flat MLP fitted best ($R^2 = 0.918$) and Screw Attention worst
($0.847$), yet Screw Attention has the highest success. Their lower success therefore does not come from a worse
fit. Their better fit is consistent with their larger function class. Without the
equivariance constraint, a network can fit regularities induced by the frame
convention of the training data. Such regularities improve the fit without improving success
in closed loop. As a result, these baselines fail under a frame change, where they retain at most
$2.7\%$ of their success. The fit to the training data therefore does not predict success in closed loop, and every result in this paper is measured by rollout success.

\begin{table}[tb]
\centering
\caption{LIBERO-Spatial: fit to the training data ($R^2$) and canonical success (\%).}
\label{tab:fit}
\footnotesize
\setlength{\tabcolsep}{3pt}
\begin{tabular}{lcc}
\toprule
Network & $R^2$ & Success \\
\midrule
Flat MLP, 438k & $0.918$ & 95.2 \\
Transformer & $0.871$ & 75.8 \\
Graph network & $0.850$ & 84.7 \\
\midrule
\textbf{Screw Attention} & $0.847$ & \textbf{97.3} \\
\bottomrule
\end{tabular}
\end{table}

\section{Where the scene relation does not help}
\label{app:nulls}

On a fixed-base manipulator, we expect the scene relation to add little in two kinds of task: reaching a goal defined by the position of one object, and avoiding obstacles with a policy that reacts to the nearest one. Because the base frame does not move, the policy needs only one vector per object, or per link, expressed in the base frame, and the scalar features already supply this information. The contribution of the relation should be more prominent when the objective is itself a relation between frames, as in the affordance task, where the policy must choose which of several candidate frames the arm can reach. Two tasks support this hypothesis.

\textbf{Reaching through clutter.} A policy given the full scene was compared with one given only the nearest obstacle to each link. With a reactive demonstrator, the difference was negligible, as expected if the nearest obstacle to each link carries the information that avoidance needs. 

\textbf{Reaching a single pose.} On the task score (success minus five times the collision rate), Screw Attention outperformed scalar only and shuffled transforms. The margin came mainly from representing the goal orientation as a screw on the end-effector token, and not from the scene relation. The object's geometric channels were zero, so permuting the object's transforms left the output of the layer unchanged. 

Whether a task depends on relations between frames can be checked before training, by regressing the demonstrator's action on object positions alone and on positions and orientations, and by testing whether more than one relation affects the action.
\FloatBarrier

\end{document}